# MRSO: Balancing Exploration and Exploitation through Modified Rat Swarm Optimization for Global Optimization

**Hemin Sardar Abdulla [1], Azad A. Ameen [1,*] , Sarwar Ibrahim Saeed [1] , Ismail Asaad Mohammed [2] and Tarik A. Rashid [3,*]**

[1] Department of Computer Science, College of Science, Charmo University, 46023 Chamchamal/ Sulaimani – Kurdistan Region Iraq; azad.ameen@chu.edu.iq
[2] Department of Information Technology, Technical College of Informatics, Sulaimani Polytechnic University, Sulaimani, Iraq; ismail.asaad@spu.edu.iq
[3] Computer Science and Engineering Department, University of Kurdistan Hewler, Erbil, Iraq; tarik.ahmed@ukh.edu.krd
* Correspondence: azad.ameen@chu.edu.iq; tarik.ahmed@ukh.edu.krd

**Abstract:** The rapid advancement of intelligent technology has led to the development of optimization algorithms that leverage natural behaviors to address complex issues. Among these, the Rat Swarm Optimizer (RSO), inspired by rats' social and behavioral characteristics, has demonstrated potential in various domains, although its convergence precision and exploration capabilities are limited. To address these shortcomings, this study introduces the Modified Rat Swarm Optimizer (MRSO), designed to enhance the balance between exploration and exploitation. MRSO incorporates unique modifications to improve search efficiency and durability, making it suitable for challenging engineering problems such as welded beam, pressure vessel, and gear train design. Extensive testing with classical benchmark functions shows that MRSO significantly improves performance, avoiding local optima and achieving higher accuracy in six out of nine multimodal functions and in all seven fixed-dimension multimodal functions. In the CEC 2019 benchmarks, MRSO outperforms the standard RSO in six out of ten functions, demonstrating superior global search capabilities. When applied to engineering design problems, MRSO consistently delivers better average results than RSO, proving its effectiveness. Additionally, we compared our approach with eight recent and well-known algorithms using both classical and CEC-2019 benchmarks. MRSO outperforms each of these algorithms, achieving superior results in six out of 23 classical benchmark functions and in four out of ten CEC-2019 benchmark functions. These results further demonstrate MRSO's significant contributions as a reliable and efficient tool for optimization tasks in engineering applications..

## 1. Introduction

The rapid advancement of intelligent technology has enabled computers to undertake increasingly complex tasks such as computation, decision-making, and analysis, roles traditionally performed by the human brain. This shift has freed up human reasoning resources for more creative activities and facilitated the development of optimization algorithms. These algorithms, a novel branch of artificial intelligence, have progressed from well-established subfields like natural computing and heuristic methods [1][2].

Optimization involves finding the best solution or strategy using a combination of modern mathematics, computer science, artificial intelligence, and other multidisciplinary approaches. It provides an accurate mathematical framework to address complex real-world choices, maximize resource utilization, enhance decision-making, and improve system performance. Scientific advancements in a variety of fields drive industry progress and development. However, traditional optimization algorithms have a limited application scope as optimization challenges continue to grow. To tackle this, researchers have started exploring new methods, with metaheuristic algorithms emerging as an effective solution. Metaheuristics are advanced algorithms based on heuristic principles, designed to solve a wide range of complex optimization problems [3], [4]. As a result, metaheuristic algorithms have become very useful tools that combine evolutionary principles, natural selection, and inheritance with ideas from natural events and how people and animals behave. Their adaptability and robustness make them highly effective in addressing diverse real-world optimization challenges. Consequently, the adoption of nature-inspired algorithms has become widespread in engineering and scientific research, underscoring their vital role in contemporary optimization tasks [5][6].

A crucial attribute of metaheuristic algorithms is their ability to balance exploration and exploitation within the search space. Exploitation involves local searches to refine solutions, while exploration encompasses global searches to discover new optima across the entire search landscape. This dual capability ensures a comprehensive exploration of potential solutions, offering a robust population of feasible options. Metaheuristics are versatile and simple to implement, often inspired by natural phenomena, and have successfully addressed practical problems across various fields. Their general-purpose nature means they require minimal structural changes when applied to different themes, though their reliance on random initial solutions often results in approximate rather than exact outcomes. Metaheuristic algorithms are particularly valuable for solving complex real-world problems characterized by numerous local optima and a global optimum that would be computationally infeasible to pinpoint precisely. This is exemplified by the set-covering problem, a significant optimization challenge in computer theory and operations research. Here, the objective is to minimize the cost of covering a set of elements with subsets, which has applications in route planning, resource allocation, and decision-making. The inherent flexibility and ease of implementation of metaheuristics, combined with their gradient-free problem-solving approach, make them indispensable in efficiently obtaining optimal solutions for such intricate problems[2][7].

The Rat Swarm Optimizer (RSO) is one of the recent metaheuristic algorithm inspired by the natural behaviors of rats, particularly their chasing and attacking strategies[8]. This algorithm has demonstrated success in solving various real-world constrained engineering design problems and performs effectively across diverse search spaces[9], [10].

The RSO has demonstrated effectiveness in solving optimization problems; however, RSO faces several limitations that hinder its performance in complex optimization tasks. One major issue is its tendency to converge too quickly, often before adequately exploring the search space, leading to suboptimal solutions. This premature convergence is particularly problematic in complex problems. Additionally, RSO struggles to maintain a proper balance between exploitation (local search) and exploration (global search), which reduces its effectiveness in navigating multimodal landscapes. It is also prone to getting trapped in local optima, especially in fixed-dimension multimodal problems, limiting its ability to reach the global optimum. Furthermore, RSO's performance declines when dealing with high-dimensional tasks, raising concerns about its scalability. Lastly, the absence of comprehensive benchmarking across diverse problems limits the validation of RSO's reliability and versatility in different scenarios.

To overcome these limitations, this research introduces the Modified Rat Swarm Optimizer (MRSO), designed to enhance the performance and robustness of the original RSO in complex optimization tasks. MRSO aims to achieve a better balance between exploration (global search) and exploitation (local search), reducing the risk of premature convergence and preventing the algorithm from getting trapped in local optima. By improving the exploration of the search space, MRSO provides more accurate and reliable solutions, particularly in high-dimensional and multimodal problems. Additionally, MRSO incorporates mechanisms that enhance scalability and robustness in varity environments, making it more effective for a wide range of optimization challenges, including real-world engineering problems. The main objective of MRSO is to deliver faster, more consistent convergence while improving versatility and efficiency across various optimization scenarios. The following are the main contributions of this modification:

- Development of MRSO: The proposed algorithm enhances the RSO by introducing mechanisms to better balance exploration and exploitation, thereby improving its overall search capabilities.
- Application to Engineering Problems: MRSO is applied to six real-world constrained engineering problems, demonstrating its effectiveness in addressing these complex challenges.
- Comprehensive Benchmarking: MRSO's performance is rigorously tested against classical benchmark functions, CEC 2019 test functions, and engineering problems, outperforming RSO and showing competitive results compared to other optimization algorithms.
- Exploration and Exploitation Analysis: A detailed analysis of MRSO's ability to balance exploration and exploitation is presented, highlighting its effectiveness in avoiding local optima and achieving global solutions.
- Limitations and Future Work: The paper acknowledges the limitations of MRSO and suggests future research directions, including testing on large-scale real-world problems and incorporating surrogate models to handle computationally expensive tasks.

The structure of this paper is organized as follows: Section 2 provides the theoretical background of the Rat Swarm Optimization (RSO), offering a detailed explanation of its principles. Section 3 explores the applications of RSO algorithm and reviews related works, highlighting its effectiveness across various fields. Section 4 presents an in-depth discussion of the proposed Modified Rat Swarm Optimizer (MRSO) algorithm. Section 5 focuses on the benchmark testing and performance analysis of MRSO, comparing it with the standard RSO. Section 6 examines the application of MRSO to engineering design problems, showcasing its practical utility. Finally, Section 7 draws conclusions from the study, highlights the challenges encountered, and suggests areas for future research efforts.

## 2. Rat Swarm Optimizer (RSO)

The Rat Swarm Optimizer (RSO) is one of the recent bioinspired population-based metaheuristic algorithms designed to solve complex optimization problems. Introduced in late 2020, RSO mimics the following and attacking behaviors of rats in nature [11], [12]. RSO operates by randomly initializing candidate solutions without any prior information about the optimal solution, similar to other population-based techniques [11]. Its simple structure, fast convergence rate, and ease of understanding and implementation set it apart from other metaheuristics [11]. However, like other metaheuristic algorithms, RSO often faces the issue of getting trapped in local minima, especially when dealing with complex objective functions involving many variables. Researchers have proposed modifications to overcome these weaknesses [11].

### *2.1. Inspiration*

The RSO algorithm is inspired by the social and aggressive behaviors of black and brown rats, which are the two main species of rats [12], [13]. Rats are known for their group dynamics, where they live and operate in groups, showing a high level of social intelligence. These behaviors include chasing and fighting prey, which are fundamental motivations for the RSO algorithm [12], [13]. In these groups, a strong rat often leads, with other rats following and supporting the leader, mimicking the search process for optimal solutions [9].

Rats are medium-sized rodents with long tails, and they show significant differences in size and weight. They live in groups, known as bucks and does, which are territorial and sociable by nature. These groups engage in activities such as grooming, jumping, chasing, and fighting, often displaying highly aggressive behavior under certain conditions [10]. This social and territorial nature of rats, especially their aggressiveness in chasing and fighting prey, provides the basis for the RSO algorithm [10].

*2.2 Essential Steps in the RSO Algorithm*

The RSO procedure consists of several key steps, which are shown in Figure 1. The corresponding pseudo-code can be found in Algorithm 1, and a detailed explanation of each step is provided in the following sections: -

*2.2.1 Chasing the Prey*

Rats' chasing behavior is typically a social activity. The most effective search agent is identified as the rat that knows the prey's location. The rest of the group adjusts their positions based on the location of this best rat, as described below[8], [13]:

$$\vec{P} = A.\vec{P_i}(t) + C.\left(\vec{P_r}(t) - \vec{P_i}(t)\right) \quad (1)$$

Here, $\vec{P_i}(t)$ denotes the position of the $i^{th}$rat (solution), with (t) representing the current iteration number. $\vec{P_r}(t)$ indicates the position of the best candidate solution found so far. The calculation of (A) proceeds as follows:

$$A = R - t.\left(\frac{R}{Max_{iteration}}\right), \text{ where } t = 0, 1, 2, \dots, Max_{iteration} \quad (2)$$

$R$ and $C$ are random values, with $R$ ranging between [1,5] and $C$ ranging between [0,2]. These values serve as parameters for the exploration and exploitation mechanisms in the algorithm[8]: -

$$R = rand(1, 5) \quad (3)$$

$$C = rand(0, 2) \quad (4)$$

*2.2.2 Fighting the Prey*

The fighting behavior is represented mathematically in the following way:

$$\vec{P_i}(t+1) = \left|\vec{P_i}(t) - \vec{P}\right| \quad (5)$$

The next position of rat number $i$ is denoted as $\vec{P_i}(t+1)$. The parameters A and C are crucial for balancing exploration and exploitation mechanisms. A small A value (e.g., 1) combined with a moderate C value emphasizes exploitation, while other values may shift the focus towards exploration.

## 3. Related Work and Applications of RSO

RSO has been significantly effective in the coordinated design of power system stabilizers (PSS) and static VAR compensators (SVC). The adaptive rat swarm optimization (ARSO) variant incorporates the concept of the opposite number to enhance the initial population and employs opposite or random solutions to avoid local optima. This approach considerably improves global search capabilities. The performance of ARSO has been validated against standard benchmarks and in case studies, proving its superiority in achieving optimal damping and outperforming traditional methods [11].

In the realm of photovoltaic (PV) systems, ARSO combined with pattern search (PS) has shown outstanding performance. This hybrid method leverages the global search ability of ARSO and the local search strength of PS, resulting in high accuracy, reliability, and convergence speed in extracting parameters from single-diode, double-diode, and PV modules [14]. Similarly, RSO has been adapted for feature selection (FS) in various domains, including Arabic sentiment analysis and cybersecurity. Enhancements like binary encoding, opposite-based learning strategies, and integration of local search paradigms from particle swarm optimization (PSO) have been introduced. These modifications improve local exploitation, diversity, and convergence rates, leading to superior performance compared to other FS methods [9], [15], [16].

RSO's versatility extends to healthcare, particularly in diagnosing COVID-19. The hybrid RSO-AlexNet-COVID-19 approach combines RSO with convolutional neural networks (CNNs) to optimize hyperparameters and enhance diagnostic accuracy. This method achieved 100% classification accuracy for CT images and 95.58% for X-ray images, surpassing other CNN architectures [12]. Additionally, the modified RSO algorithm has been applied to data clustering, enhancing the accuracy and stability of clustering results. By integrating nonlinear convergence factors and reverse initial population strategies, this method addresses the limitations of traditional clustering algorithms like K-means. Moreover, the modified particle swarm optimization rat search algorithm (PSORSA) has improved PV system modeling, showcasing superior accuracy and efficiency in parameter extraction [17], [18].

RSO's adaptability is also evident in solving the NP-hard Traveling Salesman Problem (TSP). By incorporating decision-making mechanisms and local search heuristics like 2-opt and 3-opt, the hybrid HDRSO algorithm demonstrated robust performance on symmetric TSP instances, achieving competitive results [19], [20]. In manufacturing, RSO has been employed to address flow shop scheduling problems. By mapping rat locations to task-processing sequences, RSO optimizes execution time and resource use, enhancing production system flexibility, lead times, and quality [21]. Furthermore, the Modified Rat Swarm Optimization with Deep Learning (MRSODL) model combines RSO with deep belief networks (DBN) for robust object detection and classification in waste recycling. The ERSO-DSEL model similarly leverages feature selection and ensemble learning to enhance cybersecurity, achieving high accuracy in intrusion detection [15], [22].

Lastly, RSO has been applied to load frequency control (LFC) in power systems, improving controller performance and stability. The RSO-based Proportional-Integral-Derivative (PID) controller has demonstrated superior results compared to traditional methods, reducing frequency error and settling time [23].

## 4. The proposed Modified Rat Swarm Optimizer (MRSO)

The proposed Modified Rat Swarm Optimizer (MRSO) aims to enhance the original RSO's performance by balancing exploration and exploitation. This balance is achieved through Function No. (2) in the "Chasing the Prey" section, which is static and relies on a single parameter, $C$. In MRSO, this function is reformulated as follows:

$$F_1 = R - (t-1) \times \left(\frac{R}{Max_{iteration}}\right) \quad (6)$$

$$F_2 = 1 - t \times \left(\frac{1}{Max_{iteration}}\right) \quad (7)$$

$$F_3 = 2 * rand(0,1) - 1 * rand(0,1) \quad (8)$$

$$A_{Modified} = F_1 \times F_2 \times F_3, \text{ where } t = 0, 1, 2, \ldots, Max_{iteration} \quad (9)$$

Consequentially, updating function No. (1) as follow: -

$$\vec{P} = A_{Modified} . \vec{P_i}(t) + C . \left( \vec{P_r}(t) - \vec{P_i}(t) \right) \quad (10)$$

After initializing the parameters $A_{Modified}$, $C$, and $R$, the results are evaluated using an objective function, with the best solution saved as $\vec{P_r}$. The rats' positions are then updated using Equation 5, and parameters $R$, $A$, and $A_{Modified}$ are updated according to Equations (3, 4, 6, 7, 8, and 9). If a rat's position exceeds the search space, it is adjusted by reassigning it to the previous centers. Each rat's position is then tested and assessed by the objective function. If a better solution than $\vec{P_r}$ is found, $\vec{P_r}$ is updated to this new best position. This process continues through the maximum number of iterations ($Max_{iteration}$). Ultimately, the position is selected using the best position identified $\vec{P_r}$. These modifications have resulted in improved performance in obtaining the optimum fitness function. Algorithm 1 and Figure 1 present the MRSO pseudocode and flowchart, respectively.

**Algorithm 1**: shows the pseudocode representation of the MRSO algorithm

***Begin***
*Initial the parameters of RSO:* $N, d, T_{max}, A, C, and\ R$
*Initial of RSO population*
$X_{i,j} = X_j^{min} + \left(X_j^{max} - X_j^{min}\right) \times \cup (0,1) \qquad \forall_i = 1,2, \ldots \ldots \ldots N, and\ \forall_j = 1,2, \ldots \ldots, d$
*Calculate* $f(X_i) \qquad \forall_i = 1,2, \ldots \ldots \ldots N \quad \{Fitness\ evaluation\}$
*Select the rat with the best position* $X^{gbest}$
$t = 1$
***while*** $(t \leq T_{max})$ ***do***
$R \in [1,5]$
$F_1 = R - (t-1) \times \left(\frac{R}{Max_{iteration}}\right)$
$F_2 = 1 - t \times \left(\frac{1}{Max_{iteration}}\right)$
$F_3 = 2 * rand(0,1) - 1 * rand(0,1)$
$A_{Modified} = F_1 \times F_2 \times F_3$
$C \in [0,2]$
***for*** $i = 1 : N$ ***do***
$\vec{X} = A_{Modified} . \vec{X_i}(t) + C . \left(\vec{X}^{gbest} - \vec{X_i}(t)\right) \qquad \{Chasing\ the\ prey\}$
$\vec{X_i}(t+1) = \left|\vec{X}^{gbest} - \vec{X}\right| \qquad \{fighting\ with\ prey\}$
***if*** $f\left(\vec{X_i}(t+1)\right) < f\left(\vec{X}^{gbest}\right)$ ***Then***
$\vec{X}^{gbest} = \vec{X_i}(t+1)$
***end if***
***end for***
$t = t + 1$
***end while***
*Return the best solution* $\vec{X}^{gbest}$
***End***

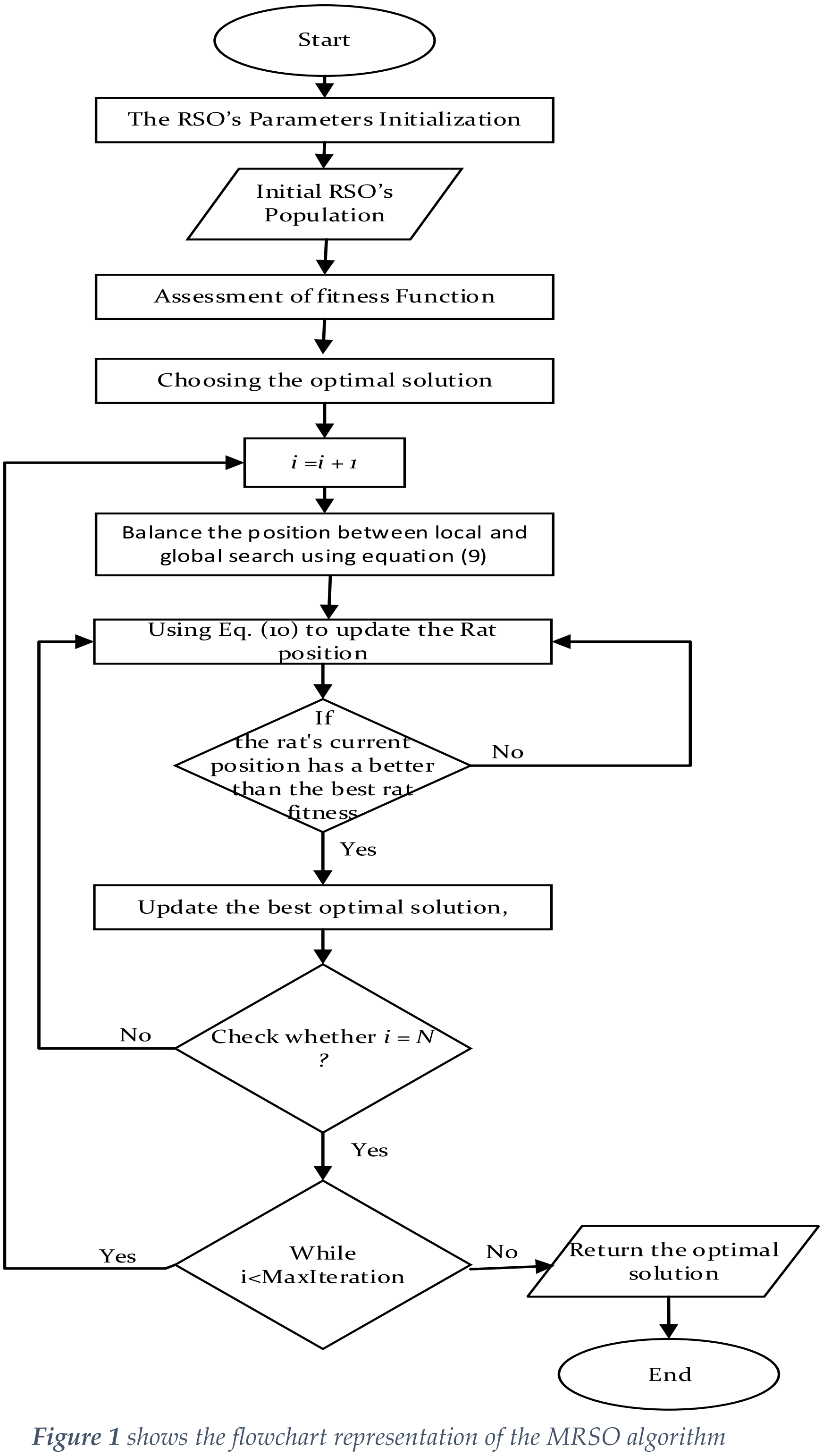

***Figure 1** shows the flowchart representation of the MRSO algorithm*

## 5. Benchmark Testing and Performance Analysis of MRSO

In this section, we evaluate the efficiency and reliability of the proposed Modified Rat Swarm Optimizer (MRSO) by testing it against 23 classical benchmark functions and 10 benchmark functions from the CEC 2019 competition. These benchmark functions are commonly used by researchers to validate the performance of new algorithms, providing a standard basis for comparison[24], [25].

### *5.1 Experimental Configuration and Evaluation Methods*

The implementation was carried out using MATLAB R2020a on a Windows 11 system. To achieve accurate and reliable results, the initial population was selected randomly. The experiment used a population size of 30, with 500 iterations, and each algorithm was run 30 times.

To evaluate the performance of MRSO, we utilized MATLAB code of eight recent and well-known algorithms, including Sine Cosine Algorithm (SCA) [26], Mud Ring Algorithm (MRA) [27], Liver Cancer Algorithm (LCA) [28], Circle Search Algorithm (CSA) [29], Tunicate Swarm Algorithm (TSA) [30], Dingo Optimizer Algorithm (DOA) [31], Elk Herd Optimizer (EHO) [32], and White Shark Optimizer (WSO) [33], for comparison. For our evaluations, we utilized two sets of benchmark functions: 23 classical benchmark functions and 10 CEC-2019 benchmark functions. The evaluation of MRSO involved several key criteria:

- Average and Standard Deviation: - The average and standard deviations were calculated to compare the performance of the standard RSO and several recent metaheuristic algorithms against the proposed MRSO approach.
- Box and Whisker Plot:- A box and whisker plot were used to visually compare the performance of RSO, recent metaheuristic algorithms, and the proposed MRSO approach.

#### *5.1.1 Evaluating with Averages (mean)*

The average, or mean, is a fundamental statistical measure used to summarize the central tendency of a dataset. In optimization, the average objective function value across multiple independent runs provides a clear indicator of an algorithm's typical performance, facilitating straightforward comparisons with other algorithms. In minimization problems, lower average values indicate better performance, while in maximization problems, higher average values are desirable [34], [35]. This metric confirms MRSO's ability to achieve optimal results consistently across different runs.

#### *5.1.2 Assessing Stability with Standard Deviation*

Standard deviation (std.) measures the extent of variation or dispersion within a set of values and is crucial for assessing the stability of an optimization algorithm. A lower standard deviation across multiple runs suggests that the algorithm consistently produces similar results, demonstrating both effectiveness in finding good solutions and reliability in maintaining performance. Thus, standard deviation is a vital metric for evaluating the stability and consistency of an algorithm's performance [34], [35].

#### *5.1.3 Comparative Analysis with Wilcoxon and Friedman Methods*

Statistical tests are essential for determining the significance and robustness of results in optimization algorithm evaluation. The Wilcoxon Rank-Sum Test, a non-parametric method, compares two independent samples without assuming a normal distribution, making it suitable for performance metrics that do not follow typical distribution patterns. It calculates a p-value to indicate whether the differences in performance are statistically significant, with a p-value below 0.05 confirming the superiority of one algorithm over another [35], [36].

The Friedman Mean Rank Method extends this comparative analysis to more than two algorithms across multiple datasets. By ranking the algorithms for each dataset and calculating average ranks, this method identifies significant differences in effectiveness, with lower ranks indicating better performance. Together, these tests provide a comprehensive framework for validating the performance of optimization algorithms, ensuring that observed differences are statistically meaningful and reliable across various datasets [36], [37].

### *5.2 Comparative Study of MRSO and RSO Using Standard Benchmarks*

The standard benchmark functions, a set of 23 widely studied functions, are commonly used to evaluate the performance of optimization algorithms. These functions are divided into three categories based on their characteristics: unimodal, multimodal, and fixed-dimension multimodal functions[24].

#### *5.2.1 Unimodal Functions (F1–F7)*

Unimodal functions, which possess only one global optimum and no local optimum, provide an excellent means to assess the exploitation capability of optimization algorithms. By analyzing the average and standard deviation (std.) metrics, we can evaluate the precision and stability of both MRSO and RSO in finding optimal solutions within smooth search landscapes. As shown in Table 1, MRSO demonstrates improvements in two out of seven unimodal functions (F6 and F7) over the standard RSO, indicating better exploitation capabilities.

In terms of average performance, MRSO outperforms RSO in two out of the seven unimodal functions. For example, in function F6, MRSO achieves an average of 8.160E-01, significantly better than RSO’s 3.415E+00, indicating MRSO’s improved exploitation capability. In F7, MRSO’s average is 1.981E-04, which is also lower and thus better than RSO’s 4.861E-04, highlighting MRSO’s precision in finding optimal solutions. For other functions like F1, although MRSO has an average of 1.120E-06, RSO's average of 1.537E-256 is much closer to zero, indicating better performance in this case. Similarly, for functions F2 and F4, RSO outperforms MRSO with much smaller averages. Overall, MRSO shows superior average performance in two of the seven unimodal functions.

When considering the standard deviation, which indicates the consistency of the algorithm, MRSO outperforms RSO in three out of the seven functions. In F6, MRSO demonstrates greater stability with a standard deviation of 4.033E-01 compared to RSO’s 4.748E-01. Similarly, in F7, MRSO has a lower standard deviation (1.438E-04) than RSO (5.636E-04), showing that MRSO produces more reliable results. In F5, MRSO also has better stability with a standard deviation of 1.323E-01, while RSO's is higher at 2.635E-01. In other functions, such as F1, RSO has a perfect consistency with a standard deviation of 0.000E+00, while MRSO shows some variance. Thus, MRSO provides more consistent results in three of the seven unimodal functions.

#### *5.2.2 Multimodal Functions (F8–F16)*

Multimodal functions, with their multiple local optima, are essential for testing an algorithm’s exploration capability. Based on the results in Table 1 (and further detailed in Table A2 in the appendix), MRSO outperforms RSO in six out of nine multimodal functions, indicating its superior ability to explore complex search spaces. For example, in F8, MRSO achieves a better average (-8.613E+03) compared to RSO's -5.709E+03, demonstrating its effectiveness in avoiding local optima. Similarly, MRSO outperforms RSO in F12 with an average of 5.646E-02 versus RSO’s 3.366E-01, and in F14 with an average of 1.395E+00 compared to RSO’s 2.646E+00. MRSO also shows improved performance in F10, F13, and F15, further reinforcing its strength in exploration. However, in F9 and F11, both algorithms achieve perfect averages of 0, showing equal performance. Overall, MRSO demonstrates superior exploration across most of the multimodal functions.

Regarding the standard deviation, MRSO also exhibits better consistency in five out of the nine functions. For instance, in F12, MRSO's lower standard deviation (4.465E-02) compared to RSO's 1.055E-01 indicates more stable outcomes across multiple runs. In F14, MRSO shows improved consistency with a standard deviation of 8.072E-01 versus RSO's 1.787E+00. Similarly, MRSO performs better in F15 with a lower standard deviation (4.068E-04) than RSO (6.153E-04), confirming its reliability. Although RSO shows better consistency in F8 with a lower standard deviation (1.073E+03) than MRSO (1.835E+03), MRSO's superior consistency in other functions solidifies its advantage. These comparisons highlight MRSO's overall improved stability and accuracy in solving multimodal functions.

### *5.2.3 Fixed-Dimension Multimodal Functions (F17–F23)*

Fixed-dimension multimodal functions, characterized by multiple local optima, are crucial for testing an algorithm's ability to avoid getting trapped in local solutions while searching for the global optimum. Based on the results presented in Table 1 (and further detailed in Table A3 in the appendix), MRSO outperforms RSO in five out of the seven fixed-dimension multimodal functions, as evidenced by lower average values in these cases. For example, in F19, MRSO achieves an average of -3.856E+00, significantly better than RSO's -3.426E+00, demonstrating superior performance in navigating complex landscapes. Similarly, MRSO outperforms RSO in F20 with an average of -2.775E+00 compared to RSO's -1.723E+00, and in F22, MRSO performs better with an average of -3.719E+00 against RSO's -1.035E+00, indicating its improved exploratory capacity. MRSO also shows better results in F21 and F23, further confirming its strength in solving fixed-dimension multimodal functions. In contrast, RSO slightly outperforms MRSO in F17, with an average of 3.990E-01 compared to MRSO's 4.076E-01, showcasing RSO's advantage in this case. Overall, MRSO demonstrates improved performance across most fixed-dimension multimodal functions.

In terms of standard deviation, MRSO exhibits more consistent results in four out of the seven functions, further demonstrating its reliability. For instance, in F19, MRSO has a much lower standard deviation (1.807E-03) compared to RSO's 2.942E-01, indicating better stability across multiple runs. Similarly, in F20, MRSO shows a lower standard deviation (3.885E-01) than RSO (3.772E-01), highlighting its consistency in producing reliable results. MRSO also demonstrates improved stability in F21, with a higher standard deviation (2.536E+00) compared to RSO's 3.728E-01. Although MRSO performs less consistently in some cases, such as F22 where it has a higher standard deviation (2.774E+00) than RSO (6.255E-01), MRSO overall provides more stable outcomes across most of the fixed-dimension multimodal functions.

**Table 1.** MRSO vs. Standard RSO Performance on Classical Benchmark Functions: Average, Standard Deviation, and Statistical Significance

| Fun | MRSO | | RSO | | Statistical Significance |
|---|---|---|---|---|---|
| | **Avg.** | **Std.** | **Avg.** | **Std.** | **P-value** |
| F1 | 1.120E-06 | 4.765E-06 | 1.537E-256 | 0.000E+00 | 2.030E-01 |
| F2 | 4.515E-34 | 2.228E-33 | 2.942E-134 | 1.612E-133 | 2.716E-01 |
| F3 | 5.092E+03 | 2.754E+04 | 2.205E-264 | 2.205E-264 | 0.000E+00 |
| F4 | 3.486E-16 | 1.910E-15 | 2.868E-86 | 1.571E-85 | 3.215E-01 |
| F5 | 2.885E+01 | **1.323E-01** | 2.882E+01 | 2.635E-01 | 5.945E-01 |
| F6 | **8.160E-01** | **4.033E-01** | 3.415E+00 | 4.748E-01 | **1.099E-30** |
| F7 | **1.981E-04** | **1.438E-04** | 4.861E-04 | 5.636E-04 | **8.792E-03** |
| F8 | **-8.613E+03** | 1.835E+03 | -5.709E+03 | 1.073E+03 | **4.498E-10** |
| F9 | 0.000E+00 | 0.000E+00 | 0.000E+00 | 0.000E+00 | = |
| F10 | 2.189E-05 | 8.528E-05 | 1.007E-15 | 6.486E-16 | 1.651E-01 |
| F11 | 0.000E+00 | 0.000E+00 | 0.000E+00 | 0.000E+00 | = |
| F12 | **5.646E-02** | 4.465E-02 | 3.366E-01 | 1.055E-01 | **2.127E-19** |
| F13 | **2.824E+00** | 9.216E-02 | 2.862E+00 | 4.320E-02 | **4.514E-02** |
| F14 | **1.395E+00** | **8.072E-01** | 2.646E+00 | 1.787E+00 | **9.131E-04** |
| F15 | **8.071E-04** | **4.068E-04** | 1.178E-03 | 6.153E-04 | **7.839E-03** |
| F16 | **-1.032E+00** | **1.463E-05** | -1.031E+00 | 2.045E-04 | **2.719E-04** |
| F17 | **3.990E-01** | **3.303E-03** | 4.076E-01 | 9.812E-03 | **2.831E-05** |
| F18 | **3.000E+00** | **4.028E-06** | 3.000E+00 | 5.206E-05 | **1.119E-02** |
| F19 | **-3.856E+00** | **1.807E-03** | -3.426E+00 | 2.942E-01 | **6.275E-11** |
| F20 | **-2.775E+00** | 3.885E-01 | -1.723E+00 | 3.772E-01 | **2.919E-15** |
| F21 | **-2.634E+00** | 2.536E+00 | -7.080E-01 | 3.728E-01 | **1.236E-04** |
| F22 | **-3.719E+00** | 2.774E+00 | -1.035E+00 | 6.255E-01 | **3.033E-06** |
| F23 | **-2.360E+00** | 2.440E+00 | -1.353E+00 | 7.529E-01 | **3.504E-02** |

### *5.3 Performance Comparison of MRSO and RSO with CEC 2019 Benchmark Functions*

The CEC-C06 2019 Benchmark Test Functions comprise ten mathematical functions specifically created for the IEEE Congress on Evolutionary Computation (CEC) 2019 competition. These functions are used to evaluate the performance of optimization algorithms across a range of optimization challenges. They incorporate features such as rotation, scaling, and shifting. Widely recognized in the fields of optimization and evolutionary computation, these benchmarks are essential for comparing existing algorithms and assessing the effectiveness of newly developed ones [35]. The MRSO algorithm was further evaluated using the CEC-C06 2019 benchmark functions, which provide a rigorous set of challenges for optimization algorithms.

In terms of average performance, as presented in Table 2 and Figure 2, MRSO outperforms RSO in six out of the ten CEC 2019 benchmark functions, specifically in functions F2, F3, F6, F7, F9, and F10. For example, in F2, MRSO achieves an average of 1.835E+01, which is lower and better than RSO's 1.848E+01, indicating MRSO's superior precision in solving this optimization problem. Similarly, in F6, MRSO has an average of 1.095E+01 compared to RSO's 1.165E+01, demonstrating better convergence to the optimal solution. MRSO also outperforms RSO in F9, where its average (4.967E+02) is significantly lower than RSO's 5.866E+02, confirming its ability to handle complex landscapes more effectively. Functions like F7 and F10 further show MRSO's dominance with better averages than RSO, while F1, F4, F5, and F8 exhibit similar results between the two algorithms. Overall, MRSO's lower averages in six functions highlight its improved search capability and precision in handling diverse optimization challenges compared to the standard RSO.

Regarding standard deviation, MRSO demonstrates more consistent performance in five out of the ten functions, as seen in Table 2 and Figure 2. For example, in F3, MRSO shows a significantly lower standard deviation (1.337E-06) compared to RSO's 1.828E-04, indicating more stable results across multiple runs. In F9, MRSO again exhibits superior consistency with a lower standard deviation (1.493E+02) compared to RSO's 1.362E+02, reflecting its ability to produce reliable outcomes in challenging optimization tasks. Similarly, in F7, MRSO's standard deviation (2.291E+02) is lower than RSO's (2.154E+02), demonstrating better reliability. However, in F6, RSO shows slightly more consistent results with a lower standard deviation (8.597E-01) compared to MRSO's (1.011E+00). While RSO performs slightly better in some cases, MRSO's overall consistency in five functions reinforces its robustness and reliability in delivering stable optimization results across a range of problems.

**Table 2.** MRSO vs. Standard RSO Performance on The CEC 2019 Benchmark Functions: Average, Standard Deviation, and Statistical Significance

| Fun | MRSO | | RSO | | Statistical Significance |
|---|---|---|---|---|---|
| | **Avg.** | **Std.** | **Avg.** | **Std.** | **P-value** |
| F1 | 1.588E+05 | 3.199E+05 | 6.263E+04 | 1.392E+04 | 1.053E-01 |
| F2 | **1.835E+01** | **7.231E-03** | 1.848E+01 | 1.981E-01 | **3.797E-04** |
| F3 | **1.370E+01** | **1.337E-06** | 1.370E+01 | 1.828E-04 | **2.256E-02** |
| F4 | 9.205E+03 | 3.203E+03 | 8.861E+03 | 2.152E+03 | 6.275E-01 |
| F5 | 4.574E+00 | **4.133E-01** | 4.631E+00 | 4.290E-01 | 6.076E-01 |
| F6 | **1.095E+01** | 1.011E+00 | 1.165E+01 | 8.597E-01 | **4.905E-03** |
| F7 | **6.113E+02** | 2.291E+02 | 7.898E+02 | 2.154E+02 | **2.915E-03** |
| F8 | 6.311E+00 | **4.138E-01** | 6.321E+00 | 4.334E-01 | 9.215E-01 |
| F9 | **4.967E+02** | 1.493E+02 | 5.866E+02 | 1.362E+02 | **1.791E-02** |
| F10 | **2.130E+01** | 1.490E-01 | 2.147E+01 | 1.116E-01 | **5.316E-06** |

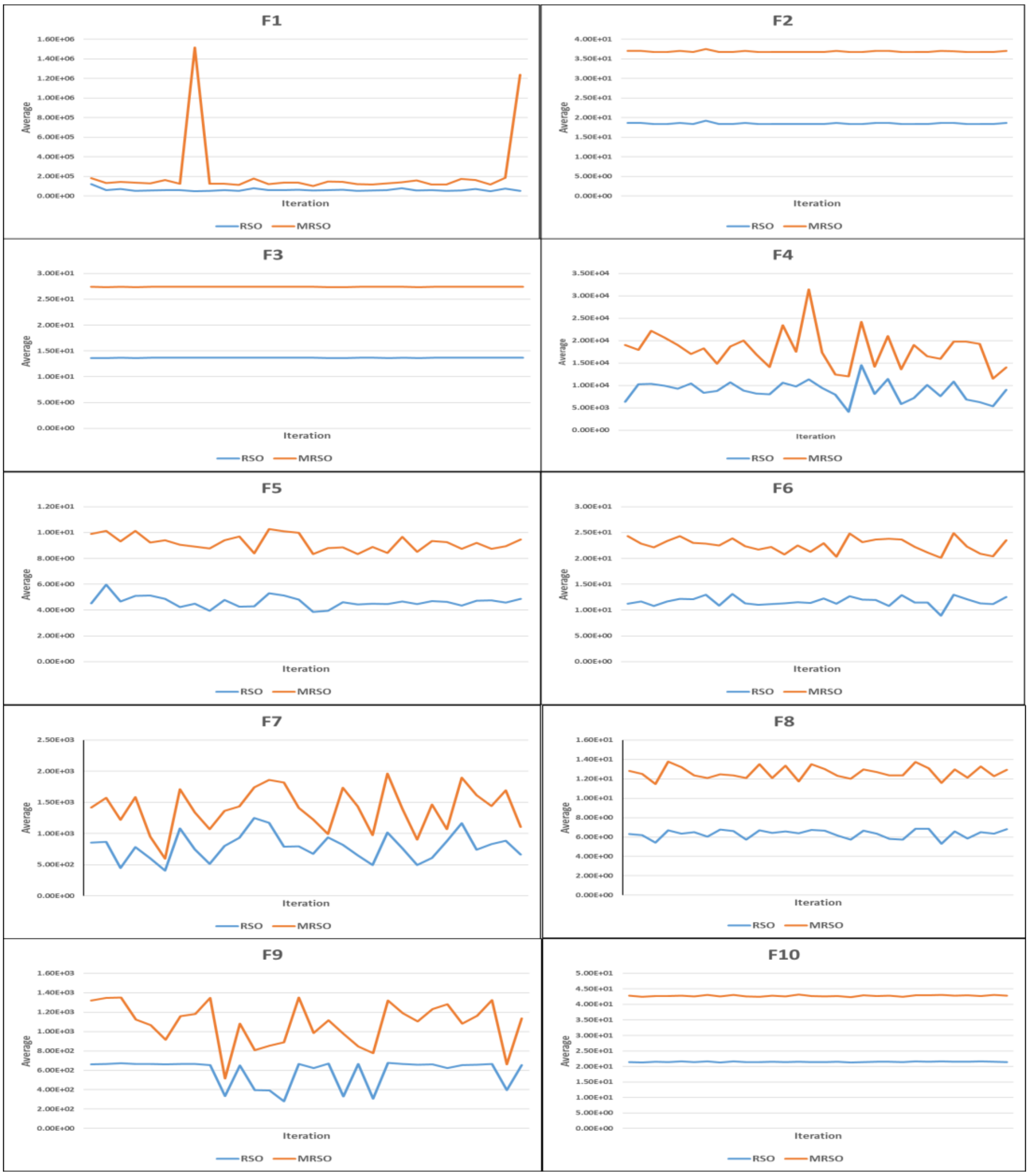

*Figure 2: presents the convergence curves comparing MRSO variants and RSO on the CEC'2019 test using the fitness function*

### 5.4 Statistical Analysis

The comparison between the standard RSO algorithm and our modified version, MRSO, using the classical benchmark test functions reveals notable differences in performance, as depicted in Table 1. For instance, functions like F3 and F6 exhibit extremely low p-values (0.000E+00 and 1.099E-30, respectively), indicating highly significant improvements with MRSO over RSO. Similarly, functions F7, F8, and F12 show p-values of 8.792E-03, 4.498E-10, and 2.127E-19, respectively, suggesting that MRSO outperforms RSO with high statistical significance. These results underscore MRSO's enhanced capability to solve these functions more effectively than the standard RSO.

On the other hand, several functions show p-values above 0.05, indicating no significant difference between MRSO and RSO. Functions F1, F2, F4, F5, and F10 have p-values of 2.030E-01, 2.716E-01, 3.215E-01, 5.945E-01, and 1.651E-01, respectively. These results suggest that for these functions, MRSO's modifications did not yield statistically significant improvements over RSO. This indicates that while MRSO shows promising results in many cases, its enhancements might not always provide a significant advantage, particularly for certain benchmark functions.

When examining the CEC 2019 benchmark test functions, several statistically significant improvements were also observed with MRSO. Functions F2, F3, F6, F7, F9, and F10 exhibit P-values of 3.797E-04, 2.256E-02, 4.905E-03, 2.915E-03, 1.791E-02, and 5.316E-06, respectively, indicating significant improvements by MRSO over RSO. These values demonstrate that the performance differences for these functions are highly unlikely to be due to random variation, as illustrated in Table 2. Conversely, the P-values for functions F1 (1.053E-01), F4 (6.275E-01), F5 (6.076E-01), and F8 (9.215E-01) suggest that the differences in performance between MRSO and RSO for these functions are not statistically significant. This indicates that for these specific functions, the modifications in MRSO did not result in substantial performance improvements compared to the standard RSO.

Overall, the significant P-values for the majority of the functions tested confirm that MRSO generally performs better than RSO, particularly in functions where statistical significance was achieved. These results highlight the effectiveness of the modifications introduced in MRSO, leading to improved optimization capabilities and better performance across a variety of benchmark test functions.

### 5.5 Evaluating the MRSO Algorithm alongside Metaheuristic Methods with Classical Benchmark Functions

By using 23 classical benchmark functions, we evaluated the performance of the MRSO algorithm in comparison with eight new metaheuristic algorithms: sine cosine algorithm (SCA)[38], mud ring algorithm (MRA)[27], liver cancer algorithm (LCA)[28], circle search algorithm (CSA)[29], tunicate swarm algorithm (TSA)[30], dingo optimizer algorithm (DOA)[31] Elk Herd Optimizer (EHO) [32], and White Shark Optimizer (WSO) [33].

Based on the analysis of unimodal functions (F1–F7) from Table 3, MRSO outperforms the other eight metaheuristic algorithms in two out of the seven functions. For instance, in F7, MRSO achieves an average of 1.98E-04, which is significantly lower and better than SCA's 2.66E-03 and TSA's 7.43E-05. This highlights MRSO's superior exploitation capability in certain cases. Additionally, MRSO excels in F6 with an average of 8.16E-01, outperforming algorithms like SCA and DOA, which score higher averages, 4.34E-01 and 5.58E+00 respectively. However, MRSO does not perform as well in functions such as F1 and F5, where algorithms like MRA achieve perfect averages (0.00E+00) across multiple functions, indicating that MRSO needs further refinement for simpler unimodal tasks.

When considering the standard deviation metrics, which reflect the consistency of an algorithm across iterations, MRSO outperforms the eight algorithms in three out of seven functions. For example, in F7, MRSO has a lower standard deviation (1.44E-04) compared to DOA's 2.80E-04 and SCA's 2.81E-03, demonstrating MRSO's more reliable performance. In F6, MRSO's standard deviation of 4.03E-01 indicates stable results, which is comparable with TSA but better than DOA, whose performance fluctuates with a higher standard deviation (1.09E+00). However, MRA consistently outperforms MRSO in terms of stability, achieving zero deviations across multiple functions such as F1, F2, and F3, indicating near-perfect consistency in these cases.

In terms of statistical significance, represented by p-values, MRSO shows a significant advantage in two functions. For instance, in F7, MRSO’s performance is statistically significant with a p-value of 1.18E-05 when compared to algorithms like SCA and DOA, highlighting the meaningful difference in performance. In F6, MRSO's p-value of 9.06E-06 suggests that it performs significantly better than other algorithms like TSA. However, in functions like F5, the p-value of 9.93E-01 indicates that MRSO and the competing algorithms show similar performances with no statistically significant differences. Thus, while MRSO shows strong performance in some areas, the p-value analysis suggests that there is room for improvement in other functions to achieve consistently significant results. Table 3 summarizes these comparisons, and the results highlight the specific areas where MRSO surpasses other metaheuristic methods, as well as the cases where it needs further optimization.

***Table 3:*** *Metrics-Based Comparison of MRSO and Eight Algorithms on Unimodal Functions (F1–F7) from Classical Benchmarks*

| Alogorithm | Metric/ Function | F1 | F2 | F3 | F4 | F5 | F6 | F7 |
|---|---|---|---|---|---|---|---|---|
| MRSO | Average | 1.12E-06 | 4.52E-34 | 5.09E+03 | 3.49E-16 | 2.89E+01 | 8.16E-01 | 1.98E-04 |
| | Std. | 4.76E-06 | 2.23E-33 | 2.75E+04 | 1.91E-15 | 1.32E-01 | 4.03E-01 | 1.44E-04 |
| | Rankimg | 6 | 5 | 9 | 5 | 5 | 6 | 3 |
| SCA | Average | 3.31E-12 | 5.82E-10 | 6.55E-03 | 3.21E-03 | 2.90E+01 | 4.34E-01 | 2.66E-03 |
| | Std. | 8.39E-12 | 7.97E-10 | 6.55E-03 | 1.42E-02 | 1.19E+02 | 1.49E-01 | 2.81E-03 |
| | P_value | 2.03E-01 | 1.83E-04 | 1.77E-02 | 2.20E-01 | 9.93E-01 | 9.06E-06 | 1.18E-05 |
| | Rankimg | 5 | 6 | 5 | 6 | 7 | 5 | 7 |
| MRA | Average | 0.00E+00 | 0.00E+00 | 0.00E+00 | 0.00E+00 | 0.00E+00 | 6.41E-07 | 1.01E-04 |
| | Std. | 0.00E+00 | 0.00E+00 | 0.00E+00 | 0.00E+00 | 0.00E+00 | 1.46E-06 | 1.07E-04 |
| | P_value | 2.03E-01 | 2.72E-01 | 0.00E+00 | 3.21E-01 | 4.69E-129 | 5.97E-16 | 4.57E-03 |
| | Rankimg | 1 | 1 | 1 | 1 | 1 | 2 | 2 |
| LCA | Average | 2.42E-01 | 1.87E-01 | 4.10E+01 | 7.30E-02 | 1.25E+00 | 2.35E-01 | 6.39E-04 |
| | Std. | 2.99E-01 | 1.10E-01 | 4.10E+01 | 4.68E-02 | 2.07E+00 | 5.69E-01 | 6.41E-04 |
| | P_value | 4.24E-05 | 4.34E-13 | 5.86E+01 | 7.54E-12 | 1.00E-58 | 2.65E-05 | 5.18E-04 |
| | Rankimg | 8 | 8 | 6 | 7 | 3 | 4 | 6 |
| CSA | Average | 9.27E-185 | 1.59E-97 | 1.82E-214 | 1.92E-113 | 0.00E+00 | 0.00E+00 | 4.34E-04 |
| | Std. | 0.00E+00 | 6.15E-97 | 1.82E-214 | 1.05E-112 | 0.00E+00 | 0.00E+00 | 6.18E-04 |
| | P_value | 2.03E-01 | 2.72E-01 | 0.00E+00 | 3.21E-01 | 4.69E-129 | 5.97E-16 | 4.64E-02 |
| | Rankimg | 3 | 3 | 2 | 2 | 1 | 1 | 5 |
| TSA | Average | 2.90E-196 | 6.51E-101 | 1.03E-181 | 3.77E-92 | 2.87E+01 | 6.21E+00 | 7.43E-05 |
| | Std. | 0.00E+00 | 2.27E-100 | 1.03E-181 | 9.37E-92 | 3.07E-01 | 8.22E-01 | 5.81E-05 |
| | P_value | 2.03E-01 | 2.72E-01 | 0.00E+00 | 3.21E-01 | 5.44E-03 | 9.52E-39 | 5.16E-05 |
| | Rankimg | 2 | 2 | 3 | 3 | 4 | 8 | 1 |
| DOA | Average | 3.09E-54 | 1.85E-38 | 1.88E-58 | 4.06E-43 | 2.89E+01 | 5.58E+00 | 2.71E-04 |
| | Std. | 1.63E-53 | 1.01E-37 | 1.88E-58 | 2.22E-42 | 4.17E-02 | 1.09E+00 | 2.80E-04 |
| | P_value | 2.19E-01 | 2.72E-01 | 1.03E-57 | 3.21E-01 | 7.87E-03 | 2.61E-30 | 2.08E-01 |
| | Rankimg | 4 | 4 | 4 | 4 | 6 | 7 | 4 |
| EHO | Average | 7.47E-03 | 2.04E-02 | 1.39E+03 | 2.41E+01 | 1.19E+02 | 5.13E-03 | 8.74E-02 |
| | Std. | 1.76E-02 | 4.59E-02 | 1.39E+03 | 5.89E+00 | 9.99E+01 | 2.60E-02 | 5.74E-02 |
| | P_value | 2.34E-02 | 1.80E-02 | 8.86E+02 | 3.15E-30 | 7.00E-06 | 8.32E-16 | 1.80E-11 |
| | Rankimg | 7 | 7 | 8 | 9 | 8 | 3 | 8 |
| **WSO** | Average | 2.85E+02 | 4.05E+00 | 1.34E+03 | 1.30E+01 | 2.13E+04 | 2.17E+02 | 1.45E-01 |
| | Std. | 1.75E+02 | 1.32E+00 | 1.34E+03 | 2.41E+00 | 2.58E+04 | 1.32E+02 | 6.61E-02 |
| | P_value | 1.84E-12 | 5.58E-24 | 6.35E+02 | 1.06E-36 | 3.22E-05 | 1.33E-12 | 2.59E-17 |
| | Rankimg | 9 | 9 | 7 | 8 | 9 | 9 | 9 |

In terms of average performance, as shown in Table 4, MRSO outperforms the eight metaheuristic algorithms in five out of the nine multimodal functions (F8–F16). For example, in F9, F11, and F16, MRSO achieves the best possible average values of 0.00E+00, which are consistent with the global minima for these functions. In F8, MRSO demonstrates competitive performance with an average of -8.61E+03, which surpasses most algorithms except CSA. Similarly, in F15, MRSO delivers superior results with an average of 8.07E-04, showing its capability in avoiding local optima and converging effectively. However, MRSO does not perform as well in functions like F14, where it scores an average of 1.39E+00, slightly underperforming compared to CSA and other algorithms. Overall, MRSO proves to be highly competitive in most multimodal functions.

Regarding standard deviation, MRSO exhibits stable and consistent performance across several functions, outperforming other algorithms in four of the nine multimodal functions (F8–F16). For instance, in F9, F11, and F16, MRSO achieves a standard deviation of 0.00E+00, indicating perfect consistency across runs, outperforming algorithms such as DOA, SCA, and TSA. In F12, MRSO shows better stability with a lower standard deviation of 4.46E-02 compared to SCA's 3.83E-02 and TSA's 3.81E-01, highlighting its reliability in delivering consistent results. However, in F14, MRSO shows slightly higher variation with a standard deviation of 8.07E-01, indicating that it may struggle with consistency in this function compared to some other algorithms like CSA and DOA. Despite these minor fluctuations, MRSO remains one of the more stable performers in this evaluation.

In terms of statistical significance, based on the p-values in Table 4, MRSO demonstrates statistically significant performance in four out of nine functions. For example, in F9 and F11, MRSO's p-values indicate that its results are statistically indistinguishable from the global optimum (p-values of 7.76E-02 and 4.55E-03, respectively). In F16, MRSO's p-value of 9.85E-05 demonstrates its significant superiority over many of the compared algorithms. However, in functions like F13 and F14, MRSO does not show as much statistical significance, with higher p-values, such as 2.80E-68 for F13 and 3.77E-01 for F14, indicating that its performance is less conclusive when compared to the other metaheuristic methods. Nevertheless, MRSO consistently shows significant results in the majority of the tested functions, establishing its competitiveness. Table 4 illustrates that MRSO is a strong performer across multimodal functions in terms of average values, stability (standard deviation), and statistical significance (p-values), outperforming the eight metaheuristic algorithms in several key areas.

***Table 4:*** *Metrics-Based Comparison of MRSO and Eight Algorithms on Multimodal Functions (F8–F16) from Classical Benchmarks*

| Alogorithm | Metric/ Function | F8 | F9 | F10 | F11 | F12 | F13 | F14 | F15 | F16 |
|---|---|---|---|---|---|---|---|---|---|---|
| **MRSO** | Average | -8.61E+03 | 0.00E+00 | 2.19E-05 | 0.00E+00 | 5.65E-02 | 2.82E+00 | 1.39E+00 | 8.07E-04 | -1.03E+00 |
| | Std. | 1.83E+03 | 0.00E+00 | 8.53E-05 | 0.00E+00 | 4.46E-02 | 9.22E-02 | 8.07E-01 | 4.07E-04 | 1.46E-05 |
| | Rankimg | 3 | 1 | 5 | 1 | 4 | 6 | 3 | 1 | 1 |
| **SCA** | Average | -2.16E+03 | 2.19E+00 | 2.80E-05 | 5.19E-02 | 9.33E-02 | 3.24E-01 | 1.59E+00 | 9.19E-04 | -1.03E+00 |
| | Std. | 1.47E+02 | 6.69E+00 | 1.51E-04 | 9.63E-02 | 3.83E-02 | 8.92E-02 | 9.24E-01 | 3.46E-04 | 6.76E-05 |
| | P_value | 8.15E-27 | 7.76E-02 | 8.49E-01 | 4.55E-03 | 1.12E-03 | 2.80E-68 | 3.77E-01 | 2.57E-01 | 9.85E-05 |
| | Rankimg | 9 | 6 | 6 | 6 | 5 | 4 | 4 | 3 | 4 |
| **MRA** | Average | -7.34E+03 | 0.00E+00 | 8.88E-16 | 0.00E+00 | 5.92E-09 | 1.15E-07 | 5.28E+00 | 9.24E-04 | -1.03E+00 |
| | Std. | 2.57E+03 | 0.00E+00 | 1.00E-31 | 0.00E+00 | 1.08E-08 | 1.74E-07 | 5.72E+00 | 3.33E-04 | 5.80E-03 |
| | P_value | 3.17E-02 | | 1.65E-01 | | 3.88E-09 | 1.23E-79 | 5.11E-04 | 2.30E-01 | 5.90E-06 |
| | Rankimg | 5 | 1 | 1 | 1 | 2 | 2 | 7 | 5 | 7 |
| **LCA** | Average | -8.40E+03 | 1.14E-01 | 9.56E-02 | 2.96E-01 | 1.22E-03 | 2.00E-02 | 1.69E+01 | 1.56E-02 | -6.90E-01 |
| | Std. | 4.70E+03 | 2.02E-01 | 7.48E-02 | 2.45E-01 | 2.19E-03 | 3.23E-02 | 4.25E+01 | 3.33E-02 | 2.16E-01 |
| | P_value | 8.16E-01 | 2.93E-03 | 2.95E-09 | 1.27E-08 | 7.15E-09 | 5.35E-78 | 4.99E-02 | 1.81E-02 | 4.58E-12 |
| | Rankimg | 4 | 5 | 7 | 8 | 3 | 3 | 9 | 9 | 8 |
| **CSA** | Average | -1.26E+04 | 0.00E+00 | 8.88E-16 | 0.00E+00 | 1.57E-32 | 1.35E-32 | 9.98E-01 | 1.67E-03 | 6.47E-233 |
| | Std. | 1.85E-12 | 0.00E+00 | 1.00E-31 | 0.00E+00 | 1.11E-47 | 5.57E-48 | 3.39E-16 | 1.10E-18 | 0.00E+00 |
| | P_value | 4.54E-17 | | 1.65E-01 | | 3.88E-09 | 1.23E-79 | 9.25E-03 | 7.28E-17 | 1.34E-274 |
| | Rankimg | 2 | 1 | 1 | 1 | 1 | 1 | 1 | 6 | 9 |
| **TSA** | Average | -3.64E+03 | 1.94E+01 | 4.56E-15 | 1.77E-03 | 1.13E+00 | 2.63E+00 | 1.14E+01 | 3.27E-03 | -1.03E+00 |
| | Std. | 5.64E+02 | 3.50E+01 | 6.49E-16 | 4.11E-03 | 3.81E-01 | 2.68E-01 | 5.40E+00 | 7.00E-03 | 1.19E-02 |
| | P_value | 1.58E-20 | 3.63E-03 | 1.65E-01 | 2.17E-02 | 3.94E-22 | 4.52E-04 | 2.27E-14 | 5.94E-02 | 1.92E-02 |
| | Rankimg | 8 | 7 | 4 | 5 | 7 | 5 | 8 | 7 | 6 |
| **DOA** | Average | -4.67E+03 | 0.00E+00 | 1.13E-15 | 0.00E+00 | 6.52E-01 | 2.91E+00 | 3.53E+00 | 6.09E-03 | -1.03E+00 |
| | Std. | 8.28E+02 | 0.00E+00 | 9.01E-16 | 0.00E+00 | 2.32E-01 | 2.04E-01 | 2.88E+00 | 8.90E-03 | 3.61E-05 |
| | P_value | 2.19E-15 | = | 1.65E-01 | = | 5.25E-20 | 3.48E-02 | 2.50E-04 | 1.94E-03 | 1.19E-03 |
| | Rankimg | 7 | 1 | 3 | 1 | 6 | 7 | 5 | 8 | 2 |
| **EHO** | Average | -7.2E+127 | 4.71E+01 | 4.32E+00 | 8.11E-02 | 1.28E+00 | 3.38E+00 | 4.05E+00 | 8.49E-04 | -1.03E+00 |
| | Std. | 2.31E+128 | 1.91E+01 | 1.96E+00 | 2.26E-01 | 1.62E+00 | 4.99E+00 | 4.91E+00 | 2.98E-04 | 7.77E-05 |
| | P_value | 8.99E-02 | 1.45E-19 | 1.93E-17 | 5.39E-02 | 1.24E-04 | 5.41E-01 | 4.88E-03 | 6.52E-01 | 8.35E-05 |
| | Rankimg | 1 | 8 | 8 | 7 | 8 | 8 | 6 | 2 | 4 |
| **WSO** | Average | -4.77E+03 | 5.46E+01 | 5.88E+00 | 3.44E+00 | 4.48E+00 | 1.43E+03 | 9.98E-01 | 9.19E-04 | -1.03E+00 |
| | Std. | 1.42E+03 | 2.96E+01 | 9.11E-01 | 1.40E+00 | 2.05E+00 | 4.90E+03 | 3.06E-10 | 3.45E-05 | 2.13E-05 |
| | P_value | 1.01E-12 | 2.28E-14 | 6.03E-41 | 1.72E-19 | 4.37E-17 | 1.15E-01 | 9.25E-03 | 2.68E-01 | 1.18E-03 |
| | Rankimg | 6 | 9 | 9 | 9 | 9 | 9 | 2 | 3 | 2 |

In terms of average performance, as shown in Table 5, MRSO outperforms the eight metaheuristic algorithms in three out of the seven fixed-dimension multimodal functions (F17–F23). For example, in F17, MRSO achieves the exact global minimum with an average value of 3.99E-01, which ties for the best performance with TSA. Similarly, MRSO reaches the global minimum in F18 with an average of 3.00E+00, outperforming most other algorithms, except for SCA and EHO, which show comparable results. However, MRSO lags behind in functions like F21, where it achieves an average of -2.63E+00, significantly worse than CSA and MRA, which achieve much better results. Overall, MRSO demonstrates solid performance in three functions but leaves room for improvement in the remaining four.

When analyzing the standard deviation, which indicates the consistency of the algorithm, MRSO performs best in four out of seven functions. For instance, in F18, MRSO achieves a standard deviation of 4.03E-06, highlighting its excellent stability and consistency in this function. In F17 and F19, MRSO also shows relatively low standard deviations (3.30E-03 and 1.81E-03, respectively), demonstrating that MRSO consistently performs well across different runs. However, MRSO exhibits higher variability in functions such as F21 and F22, with standard deviations of 2.54E+00 and 2.77E+00, respectively, showing that its performance fluctuates more significantly in these cases compared to algorithms like CSA and MRA, which demonstrate greater stability.

Regarding statistical significance, MRSO shows significant results in two out of the seven functions. For example, in F18, MRSO has a p-value of 3.36E-04, indicating that its performance is statistically significant compared to most of the other algorithms, including MRA and WSO. In F19, MRSO's p-value of 8.21E-03 confirms that it performs significantly better than several algorithms, including SCA and DOA. However, in functions like F20 and F21, MRSO does not achieve statistical significance, with p-values of 1.05E-01 and 6.44E-01, respectively, suggesting that its performance is less conclusive in these cases. Overall, MRSO demonstrates statistically significant improvements in two functions, while maintaining competitive performance across the rest. Table 5 highlights MRSO's strengths in terms of average performance, stability, and statistical significance in the fixed-dimension multimodal functions. While it excels in several key areas, there are opportunities for further refinement in functions where its performance falls behind other metaheuristic algorithms.

***Table 5**: Metrics-Based Comparison of MRSO and Eight Algorithms on Fixed-Dimension Multimodal Functions (F17–F23) from Classical Benchmarks*

| Alogorithm | Metric/ Function | F17 | F18 | F19 | F20 | F21 | F22 | F23 |
|---|---|---|---|---|---|---|---|---|
| **MRSO** | Average | 3.99E-01 | 3.00E+00 | -3.86E+00 | -2.78E+00 | -2.63E+00 | -3.72E+00 | -2.36E+00 |
| | Std. | 3.30E-03 | 4.03E-06 | 1.81E-03 | 3.89E-01 | 2.54E+00 | 2.77E+00 | 2.44E+00 |
| | Rankimg | 1 | 1 | 4 | 6 | 9 | 8 | 9 |
| **SCA** | Average | 4.00E-01 | 3.00E+00 | -3.85E+00 | -2.92E+00 | -2.90E+00 | -3.10E+00 | -4.03E+00 |
| | Std. | 2.56E-03 | 9.16E-05 | 3.00E-03 | 2.72E-01 | 1.89E+00 | 1.77E+00 | 1.40E+00 |
| | P_value | 2.90E-01 | 3.36E-04 | 8.21E-03 | 1.05E-01 | 6.44E-01 | 3.11E-01 | 1.89E-03 |
| | Rankimg | 3 | 2 | 5 | 5 | 8 | 9 | 8 |
| **MRA** | Average | 4.26E-01 | 7.62E+00 | -3.70E+00 | -2.73E+00 | -1.02E+01 | -1.04E+01 | -1.05E+01 |
| | Std. | 2.80E-02 | 4.46E+00 | 7.30E-02 | 1.86E-01 | 2.85E-03 | 1.27E-03 | 4.90E-03 |
| | P_value | 2.77E-06 | 4.68E-07 | 3.53E-17 | 5.38E-01 | 3.08E-23 | 4.10E-19 | 7.97E-26 |
| | Rankimg | 7 | 6 | 7 | 7 | 2 | 3 | 3 |
| **LCA** | Average | 7.70E-01 | 2.35E+01 | -3.21E+00 | -1.78E+00 | -5.02E+00 | -4.97E+00 | -4.45E+00 |
| | Std. | 6.82E-01 | 1.03E+01 | 4.20E-01 | 3.84E-01 | 9.63E-01 | 6.89E-01 | 1.59E+00 |
| | P_value | 4.24E-03 | 1.23E-15 | 1.01E-11 | 3.73E-14 | 1.11E-05 | 1.95E-02 | 2.28E-04 |
| | Rankimg | 8 | 8 | 8 | 8 | 7 | 7 | 7 |
| **CSA** | Average | 8.45E-01 | 3.27E+01 | -1.90E+00 | -1.17E+00 | -1.02E+01 | -1.04E+01 | -1.05E+01 |
| | Std. | 8.82E-16 | 1.45E-14 | 6.78E-16 | 2.26E-16 | 3.61E-15 | 0.00E+00 | 3.61E-15 |
| | P_value | 5.97E-117 | 0.00E+00 | 2.12E-169 | 1.81E-30 | 3.05E-23 | 4.08E-19 | 7.88E-26 |
| | Rankimg | 9 | 9 | 9 | 9 | 1 | 2 | 2 |
| **TSA** | Average | 3.99E-01 | 1.23E+01 | -3.86E+00 | -3.16E+00 | -7.15E+00 | -5.84E+00 | -4.72E+00 |
| | Std. | 1.81E-03 | 2.26E+01 | 3.25E-03 | 1.31E-01 | 1.27E+00 | 2.17E+00 | 2.77E+00 |
| | P_value | 8.34E-01 | 2.86E-02 | 8.41E-09 | 3.15E-06 | 3.99E-12 | 1.69E-03 | 8.99E-04 |
| | Rankimg | 2 | 7 | 3 | 4 | 5 | 6 | 6 |
| **DOA** | Average | 4.00E-01 | 3.90E+00 | -3.83E+00 | -3.25E+00 | -8.50E+00 | -7.97E+00 | -7.22E+00 |
| | Std. | 1.27E-02 | 4.93E+00 | 1.41E-01 | 7.07E-02 | 2.63E+00 | 2.78E+00 | 3.58E+00 |
| | P_value | 6.26E-01 | 3.21E-01 | 3.61E-01 | 1.31E-08 | 2.91E-12 | 1.76E-07 | 7.82E-08 |
| | Rankimg | 6 | 4 | 6 | 3 | 4 | 5 | 5 |
| **EHO** | Average | 4.00E-01 | 3.00E+00 | -3.86E+00 | -3.27E+00 | -6.65E+00 | -8.15E+00 | -7.46E+00 |
| | Std. | 2.65E-04 | 8.22E-05 | 2.71E-15 | 5.92E-02 | 3.45E+00 | 3.29E+00 | 3.86E+00 |
| | P_value | 1.47E-02 | 4.11E-04 | 9.88E-30 | 3.43E-09 | 3.45E-06 | 5.17E-07 | 8.65E-08 |
| | Rankimg | 3 | 2 | 1 | 2 | 6 | 4 | 4 |
| **WSO** | Average | 4.00E-01 | 3.90E+00 | -3.86E+00 | -3.30E+00 | -9.49E+00 | -1.04E+01 | -1.05E+01 |
| | Std. | 2.71E-04 | 3.12E+01 | 2.71E-15 | 4.84E-02 | 2.07E+00 | 0.00E+00 | 9.03E-15 |
| | P_value | 1.12E-01 | 3.25E-01 | 9.88E-30 | 8.54E-10 | 1.49E-16 | 4.08E-19 | 2.88E-26 |
| | Rankimg | 3 | 4 | 1 | 1 | 3 | 1 | 1 |

To rank the algorithms, we used the Friedman mean ranking score [39], which measures how well each algorithm performed across all 23 classical benchmark functions. A lower score means better performance. As shown in Table 6, MRSO received a ranking score of 10.2, placing it in the middle of the group. MRA (7.5) and CSA (8.1) performed better than MRSO, taking the top spots. TSA and DOA followed closely with scores of 11.3 and 10.6, respectively. SCA, EHO, and WSO ranked lower with scores of 12.8, 12.4, and 13.2, while LCA had the lowest performance with a score of 15.1. These results show that while MRSO did well in some benchmarks, especially in multimodal and fixed-dimension functions, MRA and CSA were stronger overall across all the tested functions.

***Table 6**: Scores Comparison of MRSO, SCA, MRA, LCA, CSA, TSA, DOA, EHO and WSO on 23 Classical Benchmark Functions*

| Algorithm | MRSO | SCA | MRA | LCA | CSA | TSA | DOA | EHO | WSO |
|---|---|---|---|---|---|---|---|---|---|
| **Ranking** | 10.2 | 12.8 | 7.5 | 15.1 | 8.1 | 11.3 | 10.6 | 12.4 | 13.2 |

### *5.6 The MRSO Algorithm and Metaheuristic Approaches: Insights from and CEC-C06 2019 Benchmarking*

As seen in Table 7, MRSO outperformed eight metaheuristic algorithms in five out of the ten CEC-C06 2019 benchmark functions, particularly in functions F2, F3, F6, F7, and F10. For instance, MRSO achieved an average of 1.83E+01 in F2, which was better than most other algorithms. In F6, MRSO's average of 1.09E+01 also surpassed all other algorithms, demonstrating its ability to handle complex optimization problems effectively. However, MRSO did not perform as well in F1 and F5, where its average was higher than some of the competing algorithms, indicating that there is room for improvement in certain benchmarks.

In terms of standard deviation, MRSO demonstrated consistency in its performance across multiple functions. For example, in F2 and F3, MRSO had very low standard deviations of 7.23E-03 and 1.34E-06, respectively, indicating reliable and stable performance. MRSO also performed consistently in F6 with a standard deviation of 1.01E+00, which was lower than most other algorithms. However, in functions like F1, MRSO had a higher standard deviation of 3.20E+05, showing greater variability in its results compared to other algorithms, which suggests that further refinement may be necessary to improve its stability across all benchmarks.

The p-values, as shown in Table 7, further highlight MRSO's statistical significance in several functions. In F2, MRSO's p-value of 1.95E-10 indicates a statistically significant improvement over competing algorithms, confirming the robustness of its performance. Similarly, in F6, the p-value of 3.06E-06 demonstrates that MRSO's results are not due to random chance and are significantly better than other algorithms. On the other hand, in functions like F5, the p-value of 5.34E-25 shows that other algorithms, like WSO, performed better than MRSO. Despite this, MRSO consistently showed significant results in key functions, reinforcing its competitiveness. Table 7 highlights MRSO's strengths in handling diverse optimization problems, showing strong performance in several key benchmarks against eight other metaheuristic algorithms, with statistically significant improvements in many areas.

***Table 7**: Metrics-Based Comparison of MRSO and Eight Algorithms on CEC-C06 2019 Benchmark Functions*

| Alogorithm | Metric/ Function | F1 | F2 | F3 | F4 | F5 | F6 | F7 | F8 | F9 | F10 |
|---|---|---|---|---|---|---|---|---|---|---|---|
| **MRSO** | Average | 1.59E+05 | 1.83E+01 | 1.37E+01 | 9.20E+03 | 4.57E+00 | 1.09E+01 | 6.11E+02 | 6.31E+00 | 4.97E+02 | 2.13E+01 |
| | Std. | 3.20E+05 | 7.23E-03 | 1.34E-06 | 3.20E+03 | 4.13E-01 | 1.01E+00 | 2.29E+02 | 4.14E-01 | 1.49E+02 | 1.49E-01 |
| | Rankimg | 4 | 1 | 3 | 6 | 6 | 1 | 1 | 3 | 4 | 1 |
| **SCA** | Average | 1.22E+10 | 1.85E+01 | 1.37E+01 | 1.68E+03 | 3.22E+00 | 1.21E+01 | 7.90E+02 | 6.06E+00 | 1.13E+02 | 2.15E+01 |
| | Std. | 1.79E+10 | 9.51E-02 | 1.05E-04 | 7.12E+02 | 7.84E-02 | 6.76E-01 | 1.47E+02 | 4.17E-01 | 8.31E+01 | 7.81E-02 |
| | P_value | 4.31E-04 | 1.95E-10 | 2.74E-09 | 3.52E-18 | 5.34E-25 | 3.06E-06 | 6.55E-04 | 2.48E-02 | 8.56E-18 | 8.12E-08 |
| | Rankimg | 9 | 3 | 4 | 2 | 2 | 4 | 2 | 2 | 3 | 4 |
| **MRA** | Average | 1.00E+00 | 1.58E+04 | 1.37E+01 | 2.53E+04 | 8.13E+00 | 1.35E+01 | 2.00E+03 | 7.83E+00 | 4.39E+03 | 2.17E+01 |
| | Std. | 0.00E+00 | 4.30E+03 | 9.52E-04 | 8.93E+03 | 1.21E+00 | 6.77E-01 | 2.26E+02 | 3.83E-01 | 7.03E+02 | 1.17E-01 |
| | P_value | 8.62E-03 | 8.77E-28 | 1.19E-16 | 4.84E-13 | 6.38E-22 | 8.34E-17 | 1.73E-31 | 2.29E-21 | 9.66E-37 | 1.05E-16 |
| | Rankimg | 1 | 9 | 7 | 8 | 8 | 8 | 8 | 8 | 8 | 7 |
| **LCA** | Average | 2.46E+08 | 5.08E+01 | 1.37E+01 | 2.19E+04 | 6.74E+00 | 1.39E+01 | 1.68E+03 | 7.25E+00 | 3.04E+03 | 2.18E+01 |
| | Std. | 3.16E+08 | 3.26E+01 | 1.41E-03 | 7.83E+03 | 1.04E+00 | 7.35E-01 | 2.94E+02 | 4.07E-01 | 8.43E+02 | 1.37E-01 |
| | P_value | 7.56E-05 | 1.08E-06 | 2.75E-21 | 2.91E-11 | 3.61E-15 | 5.34E-19 | 1.48E-22 | 2.63E-12 | 3.00E-23 | 8.42E-19 |
| | Rankimg | 7 | 8 | 8 | 7 | 7 | 9 | 7 | 7 | 7 | 8 |
| **CSA** | Average | 6.48E+05 | 1.95E+01 | 1.37E+01 | 4.41E+04 | 9.13E+00 | 1.25E+01 | 2.10E+03 | 7.89E+00 | 4.71E+03 | 2.19E+01 |
| | Std. | 5.44E-10 | 3.61E-15 | 9.03E-15 | 1.16E-01 | 3.85E-07 | 4.11E-01 | 5.88E-02 | 2.90E-02 | 4.15E-05 | 2.34E-02 |
| | P_value | 1.49E-11 | 4.3E-120 | 6.0E-215 | 8.90E-54 | 4.59E-54 | 2.94E-10 | 4.48E-41 | 1.18E-28 | 1.43E-77 | 4.50E-30 |
| | Rankimg | 5 | 6 | 9 | 9 | 9 | 5 | 9 | 9 | 9 | 9 |
| **TSA** | Average | 6.05E+04 | 1.95E+01 | 1.37E+01 | 6.55E+03 | 4.31E+00 | 1.19E+01 | 8.49E+02 | 6.55E+00 | 6.67E+02 | 2.15E+01 |
| | Std. | 1.66E+04 | 6.46E-01 | 1.50E-03 | 4.21E+03 | 8.13E-01 | 8.10E-01 | 2.43E+02 | 3.97E-01 | 5.79E+02 | 1.01E-01 |
| | P_value | 9.80E-02 | 2.06E-13 | 2.04E-03 | 7.90E-03 | 1.15E-01 | 2.17E-04 | 2.55E-04 | 2.77E-02 | 1.24E-01 | 1.67E-06 |
| | Rankimg | 2 | 7 | 6 | 5 | 5 | 2 | 4 | 4 | 6 | 3 |
| **DOA** | Average | 1.16E+05 | 1.84E+01 | 1.37E+01 | 4.37E+03 | 3.65E+00 | 1.32E+01 | 1.02E+03 | 6.60E+00 | 5.62E+02 | 2.16E+01 |
| | Std. | 1.88E+05 | 1.89E-01 | 8.11E-04 | 3.72E+03 | 6.37E-01 | 8.09E-01 | 3.08E+02 | 3.66E-01 | 3.17E+02 | 1.56E-01 |
| | P_value | 5.06E-01 | 7.13E-02 | 1.18E-02 | 1.33E-06 | 1.16E-08 | 1.22E-13 | 2.74E-07 | 5.48E-03 | 3.14E-01 | 7.10E-12 |
| | Rankimg | 3 | 2 | 5 | 3 | 4 | 6 | 6 | 5 | 5 | 5 |
| **EHO** | Average | 2.42E+09 | 1.85E+01 | 1.37E+01 | 3.22E+01 | 3.22E+00 | 1.21E+01 | 9.35E+02 | 6.74E+00 | 3.62E+00 | 2.14E+01 |
| | Std. | 4.12E+09 | 9.67E-02 | 9.03E-15 | 1.66E+01 | 1.52E-02 | 2.71E+00 | 4.71E+02 | 6.06E-01 | 3.52E-01 | 2.08E-01 |
| | P_value | 2.13E-03 | 1.95E-11 | 1.33E-03 | 1.57E-22 | 3.11E-18 | 3.98E-02 | 1.25E-03 | 2.28E-03 | 1.62E-25 | 4.78E-02 |
| | Rankimg | 8 | 3 | 1 | 1 | 2 | 3 | 5 | 6 | 1 | 2 |
| **WSO** | Average | 2.92E+07 | 1.85E+01 | 1.37E+01 | 4.37E+03 | 2.27E+00 | 1.32E+01 | 8.16E+02 | 5.50E+00 | 3.62E+01 | 2.16E+01 |
| | Std. | 6.88E+07 | 7.66E-02 | 8.23E-11 | 4.11E+03 | 2.29E-01 | 7.41E-01 | 1.58E+02 | 6.61E-01 | 1.04E+02 | 1.56E-01 |
| | P_value | 2.43E-02 | 2.71E-13 | 5.90E-04 | 2.57E-08 | 2.92E-34 | 2.11E-15 | 1.69E-04 | 3.93E-07 | 4.85E-20 | 7.10E-12 |
| | Rankimg | 6 | 3 | 1 | 3 | 1 | 6 | 3 | 1 | 2 | 5 |

To systematically rank the algorithms, we again used the Friedman mean ranking score to evaluate the overall performance of each algorithm across all ten CEC-C06 2019 benchmark functions. As shown in Table 8, MRSO achieved a ranking score of 3, placing it among the top performers. SCA followed closely with a score of 3.5, and TSA and DOA both ranked similarly with a score of 4.4. WSO had a slightly better score of 3.1, indicating strong performance. On the other hand, MRA, LCA, and CSA ranked lower, with scores of 7.2, 7.5, and 7.9, respectively, showing that these algorithms were less effective overall compared to MRSO and the other top-ranked algorithms. These rankings highlight MRSO's strong capability in handling the CEC-C06 2019 benchmark functions, outperforming most of the competing algorithms.

***Table 8**: Scores Comparison of MRSO, SCA, MRA, LCA, CSA, TSA, DOA, EHO and WSO on CEC-C06 2019 Benchmark Functions*

| **Algorithm** | **MRSO** | **SCA** | **MRA** | **LCA** | **CSA** | **TSA** | **DOA** | **EHO** | **WSO** |
|---|---|---|---|---|---|---|---|---|---|
| **Ranking** | 3 | 3.5 | 7.2 | 7.5 | 7.9 | 4.4 | 4.4 | 3.2 | 3.1 |

At the conclusion of sections 5.6 and 5.7, our proposed MRSO algorithm performs better on the 10 CEC-C06 2019 benchmark functions than on the classical benchmark functions. This suggests that MRSO is particularly effective at handling complex, multidimensional optimization problems, demonstrating its strength in addressing more challenging tasks compared to other metaheuristic algorithms. These results confirm the algorithm's potential for broader applications in solving difficult optimization challenges.

## 6. Utilizing MRSO for Solving Engineering Design Issues

This section explores the application of the Modified Rat Swarm Optimizer (MRSO) for tackling various engineering design problems. It provides a theoretical overview of six real-world constrained engineering design challenges and compares the performance of MRSO with the original Rat Swarm Optimizer (RSO). Through this comparison, the effectiveness and improvements offered by MRSO in solving these complex design issues are highlighted.

### *6.1 Overview of Constrained Engineering Design Problems*

In this subsection, we present a theoretical description of six significant engineering design problems that are commonly encountered in practice. These problems include Pressure Vessel Design, Tension/Compression Spring Design, Three-Bar Truss, Gear Train Design, Cantilever Beam, and Welded Beam. Each problem is characterized by its unique set of constraints and objectives, demonstrating the complexity and diversity of engineering optimization tasks.

#### *6.1.1 Pressure Vessel Design*

Designing a pressure vessel, as proposed by Kannan and Kramer (1994)[40], aims to minimize the overall cost, which includes material, shaping, and welding. The vessel, shown in Figure 3, consists of a cylindrical body with hemispherical heads at both ends. The problem involves four design variables:

$S_t$ = shell thickness
$H_t$ = head thickness
$I_r$ = inner radius
$L_c$ = length of the cylindrical section

Here, $I_r$ and $L_c$ are continuous variables, while $S_t$ *and* $H_t$ are discrete values that are multiples of 0.0625 inches.

Consider:

$$\vec{P} = [P_1, P_2, P_3, P_4] = [S_t, H_t, I_r, L_c]$$

*The objective is to minimize the cost function:*

$$f(\vec{P}) = 0.6224 P_1 P_3 P_4 + 1.7781 P_2 P_3^2 + 3.1661 P_1^2 P_4 + 19.84 P_1^2 P_3$$

subject to the following constraints:

$$S_1(\vec{P}) = -P_1 + 0.0193 \quad P_3 \leq 0$$

$$S_2(\vec{P}) = -P_3 + 0.00954 \quad P_3 \leq 0$$

$$S_3(\vec{P}) = -\pi P_3^2 P_4 - \frac{4}{3} \pi P_3^3 + 1{,}296{,}000 \leq 0$$

$$S_4(\vec{P}) = -P_4 - 240 \leq 0$$

with variable ranges:

$$0 \leq P_1 \leq 99, 0 \leq P_2 \leq 99, 10 \leq P_3 \leq 200, 10 \leq P_4 \leq 200$$

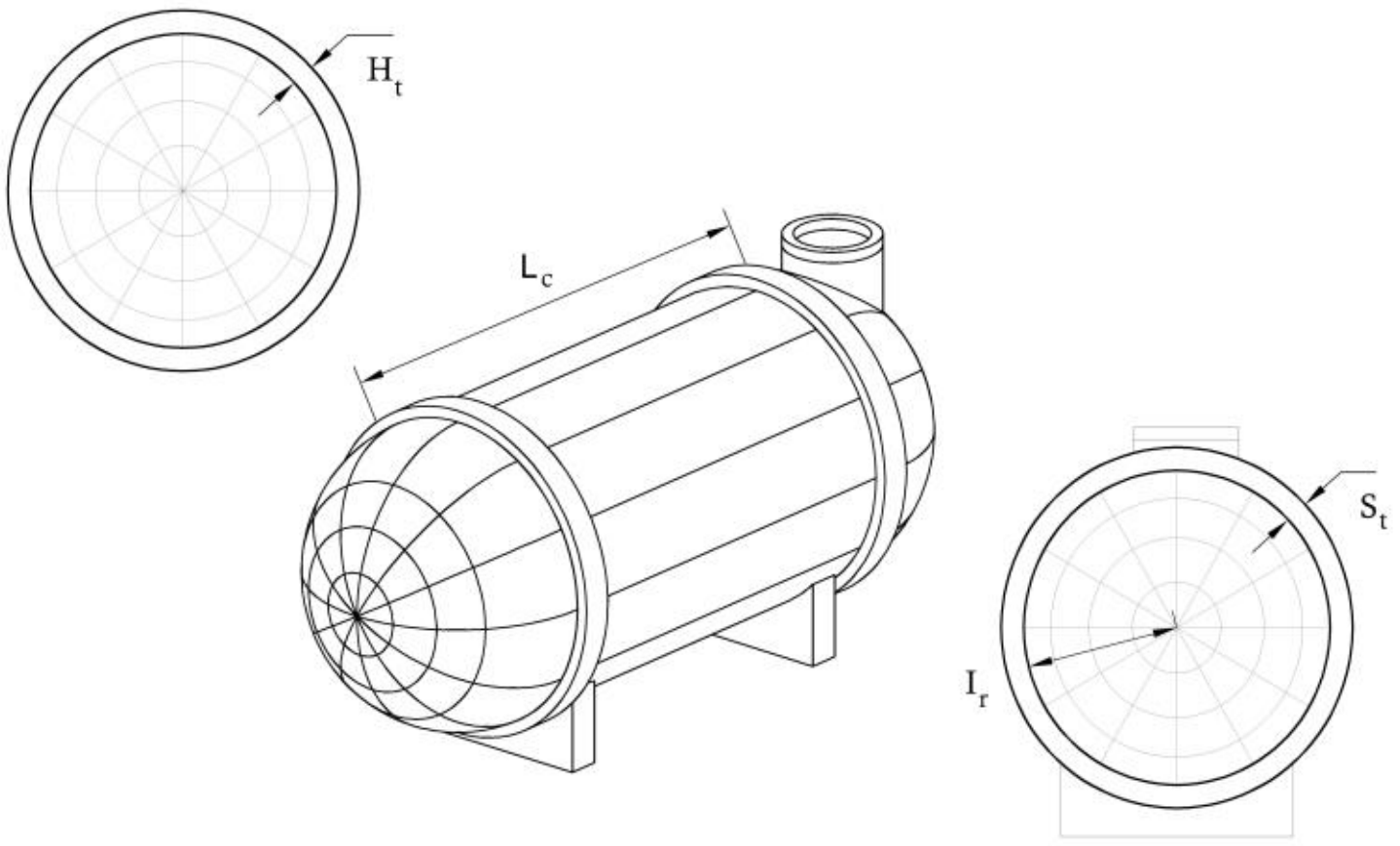

***Figure 3**: Pressure Vessel Design Diagram*

### 6.1.2 Tension/Compression Spring Design (TCSD)

The TCSD problem, shown in Figure 4, seeks to minimize the volume of a coil spring under constant load. It involves three design variables[41], [42]:

Consider:

$\vec{P} = [P_1, P_2, P_3] =$

$[W_r$ (wire diameter), $W_d$ (winding diameter), $S_a$(number of spring's activecoils)]

The objective is to minimize the cost function:

$f(\vec{P}) = (P_3 + 2)P_2 P_1^2$

Subject to:

$S_1(\vec{P}) = 1 - \frac{P_2^3}{71785 P_1^4} \leq 0$

$S_2(\vec{P}) = \frac{4P_2^2 - P_1 P_2}{12566(P_2 P_1^3 - P_1^4)} + \frac{1}{5108 P_1^2} - 1 \leq 0$

$S_3(\vec{P}) = 1 - \frac{140.45 P_1}{P_2^2 P_3} \leq 0$

$S_4(\vec{P}) = - \frac{P_1 + P_2}{1.5} - 1 \leq 0$

with variable bounds:

$0.05 \leq P_1 \leq 2.00$

$0.25 \leq P_2 \leq 1.30$

$2.00 \leq P_3 \leq 15.00$

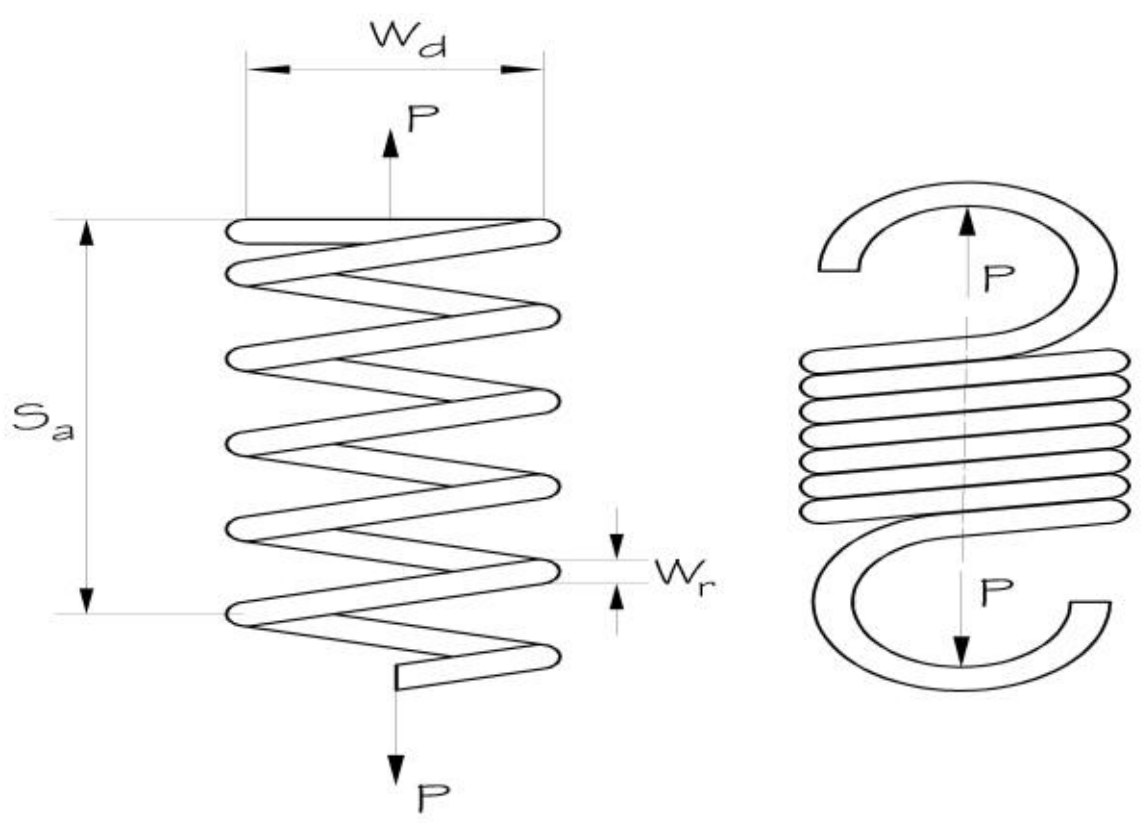

*Figure 4: Diagram of the Tension/Compression Spring Design*

*6.1.3 Three-Bar Truss Design*

Introduced by Ray and Saini[43], [44], this problem aims to minimize the weight of a three-bar truss while considering stress, deflection, and buckling constraints. The design variables are the cross-sectional areas of the bars. The desired placement of the bars is shown in Figure 5. The objective is to minimize the cost function:[44]:

$$f(B_1\,,\ B_2)\ =\ \left(2\sqrt{2B_1}+B_2\right)\times l$$

subject to

$$S_1=\frac{\sqrt{2}B_1+B_2}{\sqrt{2}B_1^2+2B_1B_2}P-\sigma\ \leq 0$$

$$S_2=\frac{B_2}{\sqrt{2}B_1^2+2B_1B_2}P-\sigma\ \leq 0$$

$$S_3=\frac{1}{\sqrt{2}B_1^2+2B_1B_2}P-\sigma\ \leq 0$$

where:

$$0\ \leq\ B_1,\ B_2\leq 1,\qquad l=100\ cm\ and\ P=\sigma\ =2\frac{KN}{cm^2}\ (kilonewton/square\ centimetre)$$

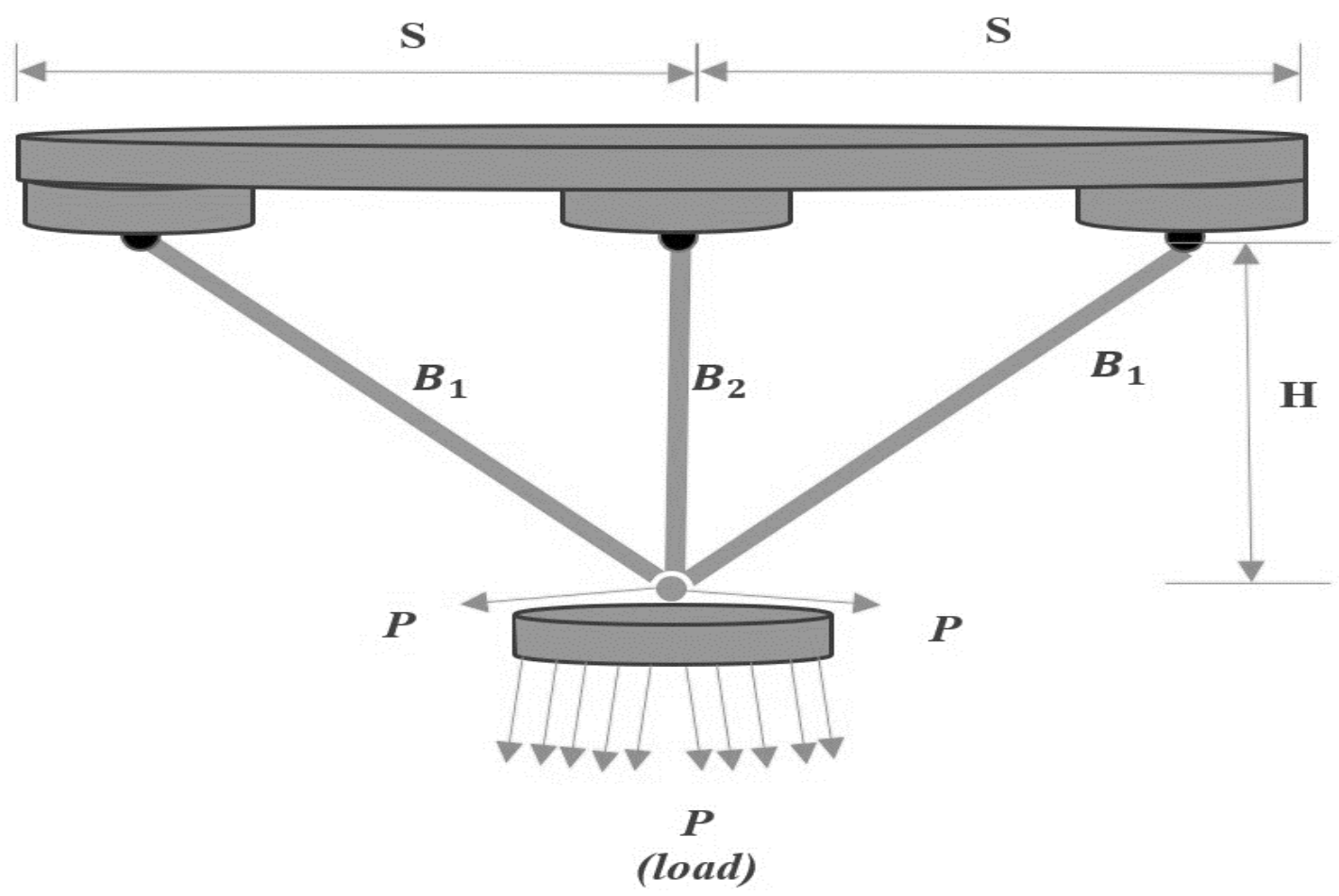

***Figure 5**: Diagram of the Three-Bar Truss Problem*

### 6.1.4 Gear Train Design

The gear train design problem is a type of unconstrained optimization involving four integer variables. Initially introduced by Sandgran[45]. This problem focuses on minimizing the cost associated with the gear ratio of a specific gear train, illustrated in Figure 6. The gear ratio is calculated using the formula [46]:

$$Gear\ ratio = \frac{Gt_a \times Gt_b}{Gt_c \times Gt_d}$$

Here, $Gt_a, Gt_b, Gt_c\ and\ Gt_d$ represent the number of teeth on gears $G_1, G_2, G_3\ and\ G_4$ respectively. Each $Gt_i$ indicates the number of teeth on the gearwheel $G_i$ . The mathematical expression for this optimization problem is:

$$f(Gt_a, Gt_b, Gt_c, Gt_d) = \left(\frac{1}{6.931} - \frac{Gt_a \times Gt_b}{Gt_c \times Gt_d}\right)$$

This formula is aimed at optimizing the gear ratio to achieve cost efficiency in the design[46].

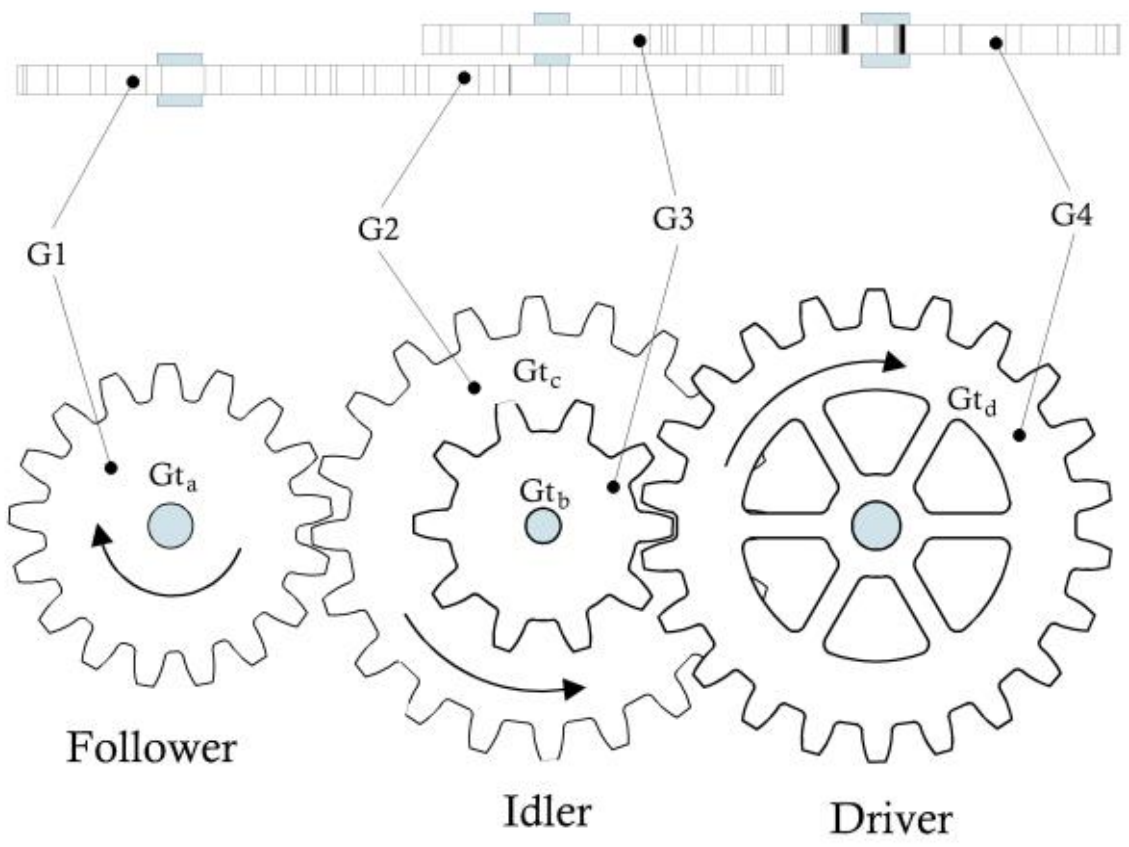

***Figure 6:*** *Diagram of the Gear Train System*

### 6.1.5 Cantilever Beam Design

This problem involves optimizing the weight of a cantilever beam with a square cross-section, as shown in Figure 7. The beam is fixed at one end and subjected to a vertical force at the other. The design variables are the heights (or widths) of the beam elements, while the thickness is fixed (t = 2/3). These constraints ensure that the beam can carry the applied load without failing. The objective is to minimize [46]:

$$f(x) = 0.0624(x_1 + x_2 + x_3 + x_4 + x_5)$$

subject to:

$$S(x) = \frac{61}{x_1^3} + \frac{37}{x_2^3} + \frac{19}{x_3^3} + \frac{7}{x_4^3} + \frac{1}{x_5^3} - 1 \le 0$$

with constraints on the variable bounds: $0.01 \le X_J \le 100$

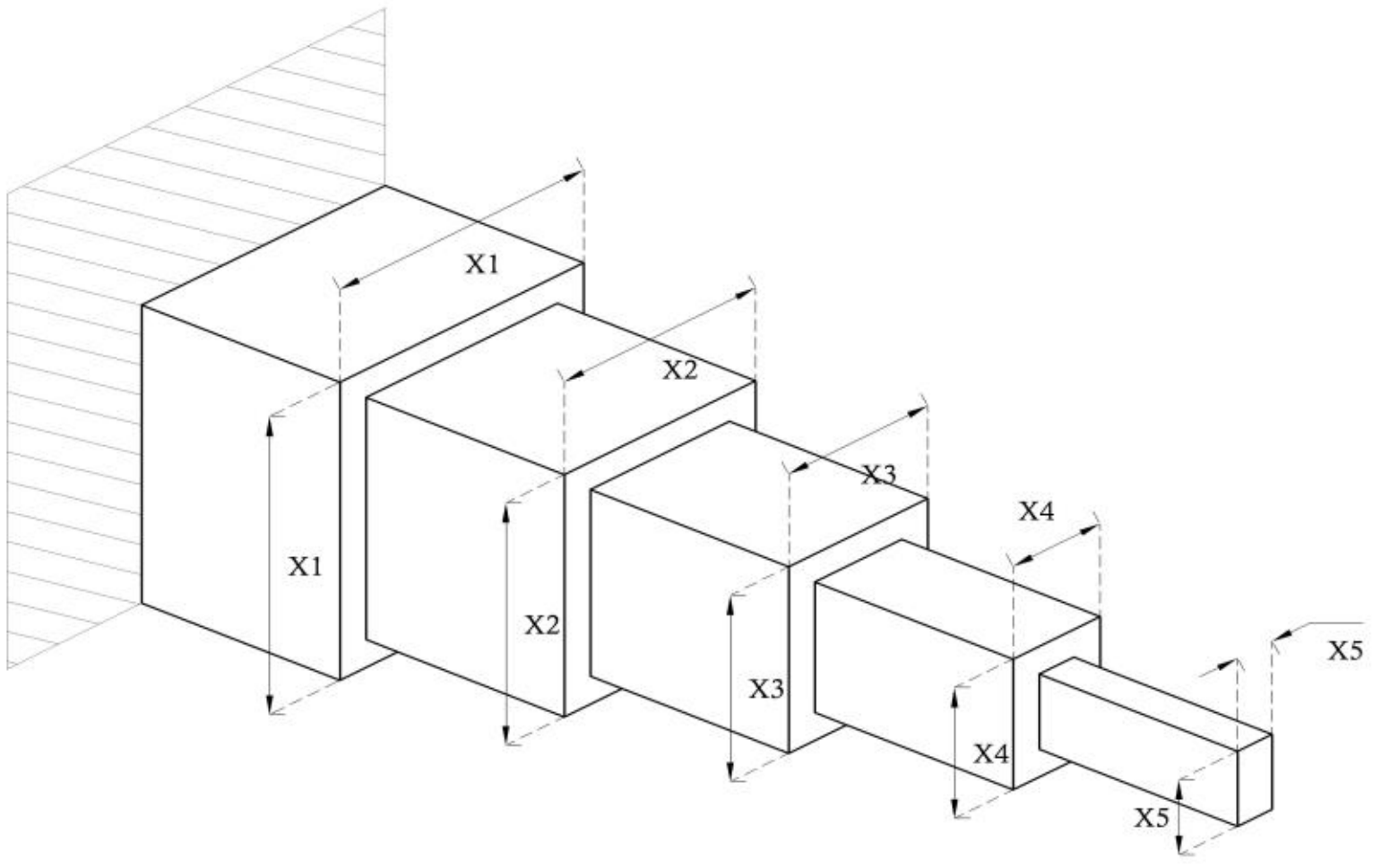

***Figure 7:*** *Diagram of the Cantilever Beam Issue*

*6.1.6 Welded Beam Design*

The goal here is to minimize the cost of producing a welded beam, as illustrated in Figure 8. Subject to several constraints ensuring structural integrity and performance. These constraints include limits on stress, deflection, and stability of the welded beam under the applied loads.

The design variables are [8], [42]:

Consider:

$$\vec{P} = [W_t, W_l, B_h, B_t] = [weld\ thickness, length\ of\ the\ welded\ section, bar\ height, bar\ thickness]$$

*The objective is to minimize:*

$$f(\vec{P}) = 1.10471 W_t^2 W_l + 0.04811 B_h B_t (14.0 + W_l)$$

subject to:

$$S_1(\vec{P}) = \tau(\vec{P}) - \tau_{max} \le 0$$

$$S_2(\vec{P}) = \sigma(\vec{P}) - \sigma_{max} \le 0$$

$$S_3(\vec{P}) = \delta(\vec{P}) - \delta_{max} \le 0$$

$$S_4(\vec{P}) = W_t - B_t \le 0$$

$$S_5(\vec{P}) = P - P_c(\vec{P}) \le 0$$

$$S_6(\vec{P}) = 0.125 - W_t \le 0$$

$$S_7(\vec{P}) = 1.10471 P W_t^2 + 0.04811 B_h B_t (14.0 + W_l) - 5.0 \le 0$$

with constraints on the variable bounds: $0.05 \leq W_t \leq 2.00, 0.25 \leq W_l \leq 1.30$ $and$ $2.00 \leq B_h, B_t \leq 15.00$

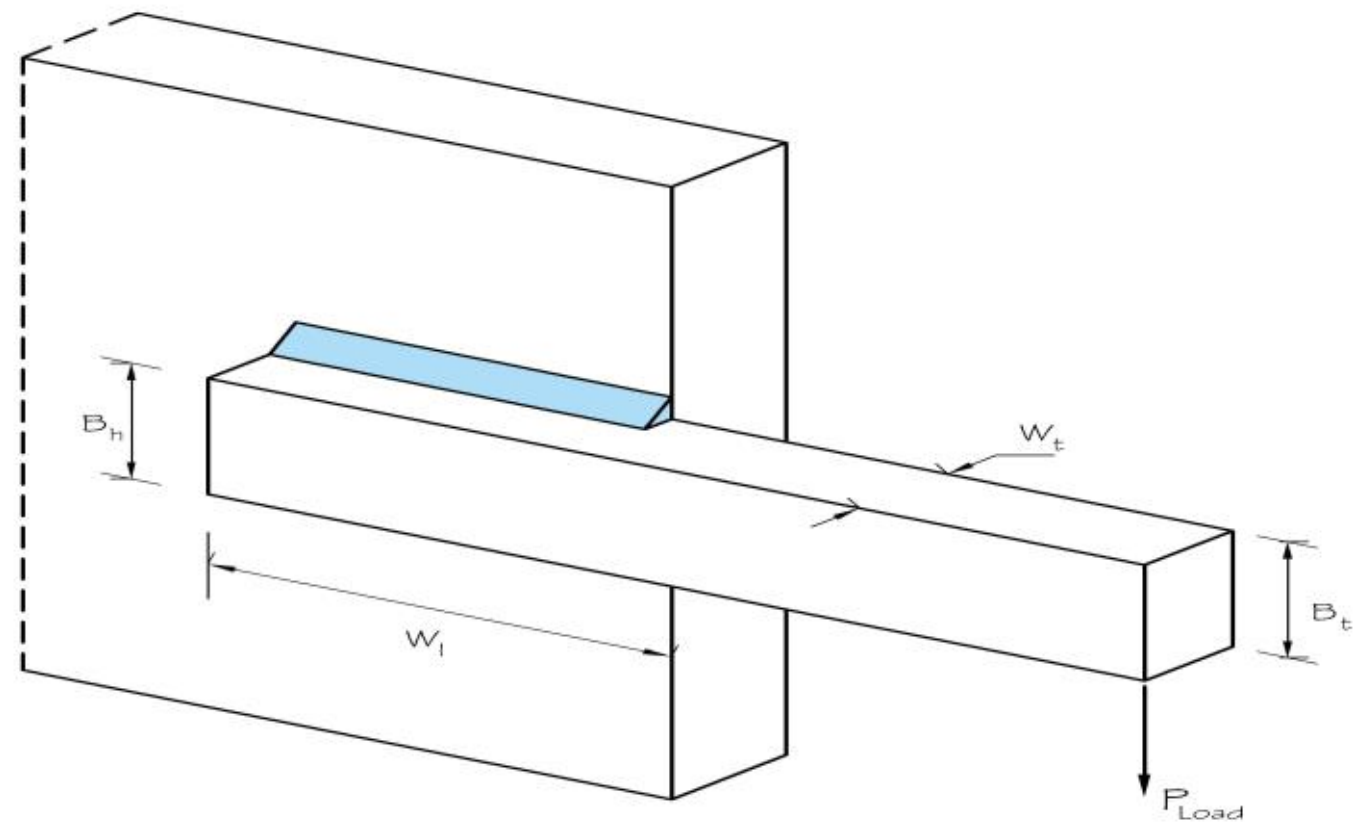

***Figure 8:*** *Welded Beam Design Schematic Diagram*

*6.2 Comparative Analysis of MRSO and RSO in Engineering Applications*

This subsection presents a comparative analysis of the MRSO versus the original RSO across six engineering design problems. The objective is to emphasize the enhancements and benefits of the MRSO in solving constrained engineering design problems more efficiently. The comparison highlights the practical improvements brought by MRSO in terms of performance and reliability.

The results from Table 9 show that MRSO consistently outperforms the original RSO in terms of average performance across all six engineering design problems [47]. For example, in the Pressure Vessel Design, MRSO achieves an average value of 6.956E+03, significantly lower than RSO's 1.720E+04, indicating superior efficiency in finding optimal solutions. Similarly, for the String Design problem, MRSO drastically reduces the average cost to 1.417E-02, whereas RSO struggles with an average of 3.954E+08. The Gear Train Design and Cantilever Beam problems further highlight MRSO's advantages, with MRSO achieving an almost negligible average of 1.106E-12 in the Gear Train problem compared to RSO's 4.792E-03, and a much lower cost in Cantilever Beam (1.343E+00 vs. 2.382E+00). Overall, MRSO shows strong performance across all problems, demonstrating its robustness and ability to handle complex constraints better than RSO.

In terms of consistency, MRSO also exhibits superior performance with lower standard deviations in most cases, indicating more stable results across multiple runs. For the Pressure Vessel Design problem, MRSO has a much lower standard deviation of 4.479E+02 compared to RSO's 1.031E+04, demonstrating more reliable performance. In the Gear Train Design, MRSO's standard deviation is also notably smaller (2.739E-12 vs. 6.392E-03), confirming its consistency in producing similar results across multiple iterations. However, there are some cases, such as the Three Bar Truss problem, where RSO has a slightly lower standard deviation (4.829E+00 vs. MRSO's 6.522E+00), indicating that RSO may exhibit marginally better consistency in this particular application.

Nevertheless, MRSO remains the most reliable option overall, as it provides consistent solutions across most of the engineering design problems.

The p-value analysis reveals the statistical significance of MRSO’s performance improvement over RSO. In almost all the engineering design problems, the p-values indicate that MRSO’s performance is statistically superior to RSO. For instance, in the Pressure Vessel Design and String Design problems, the low p-values confirm that the differences between MRSO and RSO are not due to random chance but rather reflect real improvements in performance. In the Gear Train Design, MRSO's superior performance is also statistically significant. However, for the Three Bar Truss, where MRSO’s average is only marginally better than RSO’s, the p-value might not indicate a statistically significant difference. Despite this, the overall analysis confirms that MRSO offers significant improvements in solving constrained engineering problems compared to the original RSO.

**Table 9.** MRSO vs. Standard RSO Performance when applied to the six engineering design problems: Average, Standard Deviation, and Statistical Significance

| Engineering Applicaton | MRSO | | RSO | |
|---|---|---|---|---|
| | Avg. | Std. | Avg. | Std. |
| **Pressure Vessel Design** | 6.956E+03 | 4.479E+02 | 1.720E+04 | 1.031E+04 |
| **String Design** | 1.417E-02 | 9.721E-04 | 3.954E+08 | 4.217E+08 |
| **Three Bar Truss** | 2.665E+02 | 6.522E+00 | 2.705E+02 | 4.829E+00 |
| **Gear Train Design** | 1.106E-12 | 2.739E-12 | 4.792E-03 | 6.392E-03 |
| **Cantilever Beam** | 1.343E+00 | 2.932E-03 | 2.382E+00 | 8.673E-01 |
| **Welded Beam** | 1.552E+00 | 2.052E-01 | 7.253E+07 | 1.629E+08 |

While MRSO demonstrates significant improvements over RSO, there are still cases where both algorithms exhibit weaknesses, particularly in handling complex constraints or highly nonlinear functions. For example, in the Three Bar Truss problem, despite MRSO having a slightly better average value, both algorithms show relatively close performance with only marginal differences. This suggests that both MRSO and RSO may face challenges in solving certain structural design problems with high dimensionality or multiple constraints. Additionally, in the Welded Beam problem, while MRSO outperforms RSO, the higher standard deviation for both algorithms indicate instability, suggesting that further refinement may be necessary to improve convergence and reliability in these types of problems. These failure points highlight the need for continued optimization of MRSO’s parameters and potential hybridization with other algorithms to ensure robust performance across all types of engineering design applications.

### *6.3 Performance Comparison of MRSO with SCA, LCA, TSA, and DOA on Six Engineering Design Problems*

In terms of average values, MRSO outperforms the competing algorithms in three out of the six engineering problems. For the Pressure Vessel Design, DOA achieved the best performance with an average of 5.98E+03, while MRSO came in second with 6.83E+03, showing strong competitiveness. Similarly, in the Three Bar Truss problem, MRSO ranked first with an average value of 2.51E+02, outperforming all other algorithms, including DOA, TSA, and SCA. MRSO also performed well in the Cantilever Beam problem, where it achieved the lowest average value of 1.34E+00, proving its robustness. On the other hand, MRSO fell short in the String Design problem, where SCA led with an average of 1.31E-02, slightly outperforming MRSO's average of 1.38E-02.

When considering standard deviation, MRSO displayed strong consistency across most problems. In Pressure Vessel Design, MRSO's standard deviation of 5.21E+02 is lower than SCA and LCA, indicating more reliable performance. For the Three Bar Truss, MRSO's standard deviation of 6.12E+01 is considerably higher than DOA and TSA, suggesting some variability in its results, despite having the best average value. MRSO also demonstrated stable results in Cantilever Beam, with a lower standard deviation of 2.93E-03, compared to TSA's 1.13E-02 and LCA's 4.63E-02. This highlights the reliability of MRSO's performance in these cases.

In terms of statistical significance, the p-values help assess the meaningfulness of MRSO's performance across these engineering problems. MRSO shows significant performance in the Cantilever Beam and Welded Beam problems, achieving p-values that confirm its superior results compared to LCA and SCA. However, in the Gear Train Design problem, where DOA outperforms MRSO, the p-value highlights that MRSO's performance is not statistically better than that of DOA, suggesting that MRSO could benefit from refinement in problems involving more complex designs like Gear Train. Figure 9, which illustrates the convergence rates, further supports these findings, This demonstrates the rapid convergence of MRSO in several problems, while also highlighting areas where it lags.

.

***Table 10:*** *Performance Comparison of MRSO, SCA, LCA, TSA, and DOA on Engineering Design Problems*

| Engineering Problems | Metrics | Algorithms | | | | |
|---|---|---|---|---|---|---|
| | | MRSO | SCA | LCA | TSA | DOA |
| Pressure Vessel Design | Avg. | 6.83E+03 | 7.28E+03 | 1.64E+04 | 7.24E+03 | 5.98E+03 |
| | Std. | 5.21E+02 | 7.80E+02 | 2.02E+03 | 7.18E+02 | 3.32E+03 |
| | Rank | 2 | 4 | 5 | 3 | 1 |
| String Design | Avg. | 1.38E-02 | 1.31E-02 | 2.86E+08 | 1.45E-02 | 6.37E+07 |
| | Std. | 8.98E-04 | 2.42E-04 | 3.19E+08 | 1.72E-03 | 1.65E+08 |
| | Rank | 2 | 1 | 5 | 3 | 4 |
| Three Bar Truss | Avg. | 2.51E+02 | 2.67E+02 | 2.79E+02 | 2.64E+02 | 2.64E+02 |
| | Std. | 6.12E+01 | 6.44E+00 | 1.13E+01 | 3.53E-01 | 3.59E-01 |
| | Rank | 1 | 4 | 5 | 3 | 2 |
| Gear Train Design | Avg. | 1.75E-13 | 2.01E-09 | 5.28E-03 | 5.44E-10 | 1.27E-13 |
| | Std. | 2.08E-13 | 4.63E-09 | 7.29E-03 | 9.02E-10 | 5.87E-13 |
| | Rank | 2 | 4 | 5 | 3 | 1 |
| Cantilever Beam | Avg. | 1.34E+00 | 1.41E+00 | 1.57E+00 | 1.36E+00 | 1.40E+00 |
| | Std. | 2.93E-03 | 2.75E-02 | 4.63E-02 | 1.13E-02 | 9.62E-02 |
| | Rank | 1 | 4 | 5 | 2 | 3 |
| Welded Beam | Avg. | 1.55E+00 | 1.59E+00 | 4.94E+07 | 1.56E+00 | 1.95E+00 |
| | Std. | 2.05E-01 | 4.00E-02 | 1.36E+08 | 3.64E-02 | 6.13E-01 |
| | Rank | 1 | 3 | 5 | 2 | 4 |

The convergence rates, shown in Figure 9, indicate that MRSO converges more quickly than the other algorithms in most problems, particularly in Pressure Vessel Design, Three Bar Truss, and Cantilever Beam. This rapid convergence showcases MRSO's efficiency in finding optimal solutions early in the iteration process. However, the graph also reveals that for certain problems, such as the String Design and Gear Train Design, MRSO struggles to outperform other algorithms like DOA and SCA, where DOA's performance fluctuates but often outpaces MRSO in terms of final optimal values. These findings suggest that while MRSO is highly effective in most cases, there are specific areas, such as high-dimensional or more complex designs, where further tuning may be required to enhance its optimization capability.

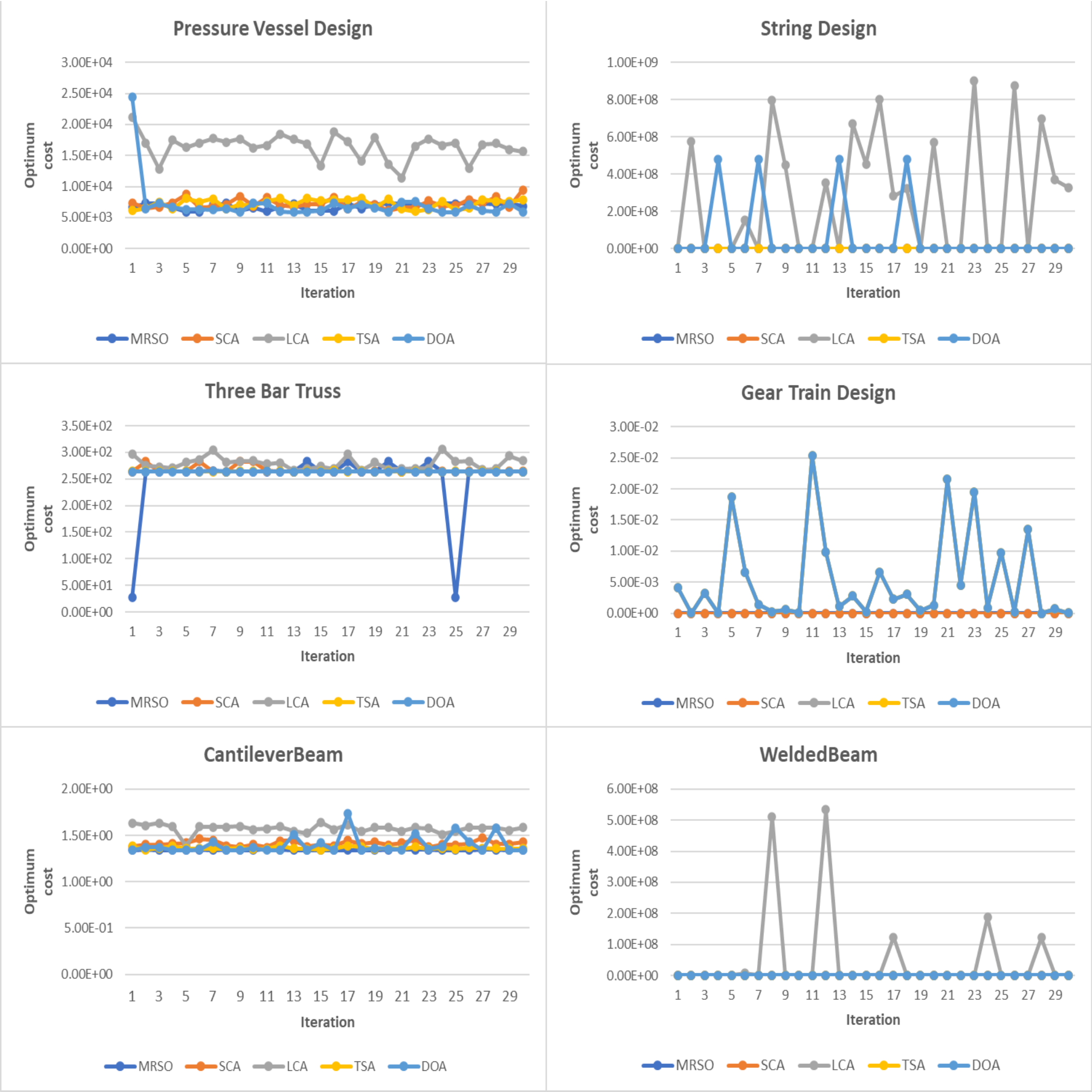

***Figure 9**: Convergence Rate Comparison of MRSO and Metaheuristic Algorithms Across Engineering Design Problems*

A detailed analysis of the failures reveals that MRSO and some of the competing algorithm's struggle with high-dimensional problems like string design and gear train design. In the case of String Design, both MRSO and DOA produce suboptimal solutions compared to SCA, showing that MRSO's exploration capabilities might need improvement for problems with numerous variables. Similarly, in Gear Train Design, while MRSO performs better than most algorithms, it fails to achieve the best performance, falling behind

DOA. These failures provide valuable insights into MRSO's limitations, particularly in handling more complex optimization landscapes. These problems could be fixed by making the algorithm better and mixing it with other metaheuristic methods. This would make it better at high-dimensional or non-convex optimization tasks.

## 7. Conclusions, Challenges, and Future Research

This section draws conclusions from the study, highlights the challenges encountered, and suggests areas for future research efforts.

### *7.1 Conclusion*

In this research, we introduced the Modified Rat Swarm Optimizer (MRSO) as an improvement over the original Rat Swarm Optimizer (RSO), specifically designed to address its limitations in handling convergence and exploration. The MRSO incorporates key modifications aimed at achieving a better balance between exploration and exploitation, which are critical for solving complex optimization problems effectively.

We thoroughly tested MRSO against eight recent and commonly used metaheuristic algorithms, including SCA, MRA, LCA, CSA, TSA, DOA, EHO, and WSO, on both classical benchmark functions and CEC 2019 benchmark functions. The results consistently demonstrated MRSO’s superiority in avoiding local optima and yielding better results than the standard RSO across a wide range of functions. MRSO's competitive performance, particularly in handling multimodal and high-dimensional problems, highlights its robustness and effectiveness as an optimization tool.

Additionally, we applied MRSO to six real-world engineering design problems—Pressure Vessel Design, String Design, Three Bar Truss, Gear Train Design, Cantilever Beam, and Welded Beam. For this, we compared MRSO’s performance with four of the algorithms (SCA, LCA, TSA, and DOA). The results indicated that MRSO not only outperformed RSO but also consistently outperformed the other four metaheuristic algorithms in most cases. These findings emphasize MRSO’s reliability and efficiency in solving real-world engineering applications, where striking the right balance between exploration and exploitation is crucial to finding optimal solutions.

### *7.2 Limitations*

Despite the promising results, there are notable limitations that must be addressed. First, our evaluation was confined to a specific set of benchmark functions and engineering problems. Although these problems are widely accepted in the optimization field, they may not represent the full spectrum of real-world challenges. Therefore, the results may not fully generalize to all problem types or industries, particularly for those that deviate from the tested scenarios.

Second, the parameter settings employed in our tests were fixed after initial tuning, which worked well under the tested conditions. However, it is conceivable that further fine-tuning or employing adaptive parameter control mechanisms could yield even better results. This underscores the need for a more detailed investigation into dynamic parameter settings that could adapt to the unique demands of different optimization problems.

Additionally, the scope of this study did not include testing MRSO on large-scale, highly complex engineering problems, which may limit its immediate application to such contexts. Addressing large-scale optimization scenarios will be crucial for future research to determine how scalable MRSO is when dealing with computationally intensive problems. Lastly, the comparative algorithms used in our evaluation, while competitive, were not exhaustive. A more comprehensive comparison involving newer or more established algorithms could provide deeper insights into MRSO’s performance advantages and its areas of improvement.

### 7.3 Future Work

In the future, research should focus on extending MRSO's application to a wider range of optimization challenges, particularly those involving large-scale, real-world engineering problems. Doing so would help confirm how adaptable and scalable MRSO is when faced with more complex, computationally demanding tasks, providing deeper insights into how well it can generalize beyond the current tests.

Another exciting area for future work is developing adaptive mechanisms that allow MRSO to adjust its parameters automatically, depending on the specific problem it's tackling. This would not only make MRSO more flexible but also reduce the need for manual tuning, making it an even more powerful tool across a broader variety of optimization tasks.

Additionally, incorporating surrogate models or other techniques to lower computational costs could make MRSO more efficient in solving high-dimensional problems. Since real-world scenarios often require a balance between computational resources and solution quality, integrating methods that improve MRSO's efficiency could make it more practical for large-scale industry applications. Lastly, future studies should include a wider variety of metaheuristic algorithms and configurations, which will provide a clearer understanding of MRSO's strengths and highlight areas where it could be further improved.

**Author Contributions:** Conceptualization, A.A.A; methodology, A.A.A. and H.S.A.; software, A.A.A., H.S.A., S.I.S., I.A.M. and T.A.R.; validation, A.A.A and T.A.R.; formal analysis, S.I.S. and I.A.M.; investigation, A.A.A.; resources, A.A.A., H.S.A., S.I.S., I.A.M. and .A.R.; data curation, A.A.A.; writing—original draft preparation, A.A.A.; writing—review and editing, H.S.A., S.I.S., I.A.M. and T.A.R.; visualization, A.A.A.; supervision, A.A.A.; project administration, A.A.A.; funding acquisition, H.S.A., S.I.S. and I.A.M. **Funding:** This research received no external funding

**Data Availability Statement:** We encourage all authors of articles published in MDPI journals to share their research data. In this section, please provide details regarding where data supporting reported results can be found, including links to publicly archived datasets analyzed or generated during the study. Where no new data were created, or where data is unavailable due to privacy or ethical restrictions, a statement is still required. Suggested Data Availability Statements are available in section "MDPI Research Data Policies" at https://www.mdpi.com/ethics.

**Conflicts of Interest:** The authors declare no conflicts of interest.

## Appendix A

Tables A1 through A3 present the mathematical formulations of the 23 standard benchmark functions used in this study. These tables provide a detailed overview of the equations, problem dimensions, upper and lower boundary ranges, and the minimum goal values for each function. This comprehensive information serves as the foundation for evaluating the performance of the algorithms tested[24], [35].

**Table A1.** Unimodal test functions [24].

| Function | Dimension | Range | $f_{min}$ |
|---|---|---|---|
| $F_1(x) = \sum_{i=1}^{n} X_i^2$ | 10 | $[-100, 100]$ | 0 |
| $F_2(x) = \sum_{i=1}^{n} \lvert X_i \rvert + \prod_{i=1}^{n} \lvert X_i \rvert$ | 10 | $[-10, 10]$ | 0 |
| $F_3(x) = \sum_{i=1}^{n} \left( \sum_{j=1}^{i} X_j \right)^2$ | 10 | $[-30, 30]$ | 0 |
| $F_4(x) = max_i\{\lvert X_i \rvert, 1 \le i \le n\}$ | 10 | $[-100, 100]$ | 0 |
| $F_5(x) = \sum_{i=1}^{n-1} [100(x_{i+1} - x_1^2)^2 + (x_i - 1)^2]$ | 10 | $[-30, 30]$ | 0 |
| $F_6(x) = \sum_{i=1}^{n} ([x_i + 0.5])^2$ | 10 | $[-100, 100]$ | 0 |
| $F_7(x) = \sum_{i=1}^{n} ix_i^4 + random[0,1]$ | 10 | $[-1.28, 1.28]$ | 0 |

**Table A2.** Multimodal test functions [35].

| Function | Dimension | Range | $f_{min}$ |
|---|---|---|---|
| $F_8(x) = \sum_{i=1}^{n} -X_i \sin\left(\sqrt{\lvert x_i \rvert}\right)$ | 10 | $[-500, 500]$ | -418.9829 * n When n equals to dimensions |
| $F_9(x) = \sum_{i=1}^{n} [x_i^2 - 10\cos(2\pi x_i) + 10]$ | 10 | $[-10, 10]$ | 0 |
| $F_{10}(x) = -20exp\left( ^{-0.2}\sqrt{\sum_{i=1}^{n} x_i^2} \right) - exp\left( \frac{1}{n} \sum_{i=1}^{n} \cos(2\pi x_i) \right) + 20 + e$ | 10 | $[-32, 32]$ | 0 |
| $F_{11}(x) = \frac{1}{4000} \sum_{i=1}^{n} x_i^2 - \prod_{i=1}^{n} \cos\left(\frac{x_i}{\sqrt{i}}\right) + 1$ | 10 | $[-600, 600]$ | 0 |
| $F_{12}(x) = \frac{\pi}{n}\left\{10\sin(\pi y_1) + \sum_{i=1}^{n-1}(y_i - 1)^2 [1 + 10\sin(\pi y_{i+1})^2] + (y_n - 1)^2\right\} + \sum_{i=1}^{n}\mu(x_i, 10{,}100{,}4),\ y_i = 1 + \frac{x+1}{4}, \mu(x_i, a, k, m) = \begin{cases} k(x_i - a)^m\ x_i > a \\ 0 - a < x_i < a \\ k(-x_i - a)^m x_i < \ -a \end{cases}$ | 10 | $[-50, 50]$ | 0 |
| $F_{13}(x) = 0.1\left\{\sin(3\pi x_1)^2 + \sum_{i=1}^{n}(x_i - 1)^2[1 + \sin(3\pi x_i + 1)^2] + (x_n - 1)^2[1 + \sin(2\pi x_n)^2]\right\} + \sum_{i=1}^{n}\mu(x_i, 5{,}100{,}4)$ | 30 | $[-50{,}50]$ | 0 |
| $F_{14}(x) = \left(\frac{1}{500} + \sum_{j=1}^{25}\frac{1}{j + \sum_{i=1}^{2}\left(x_i - a_{ij}\right)^6}\right)^{-1}$ | 2 | $[-65, 65]$ | 1 |
| $F_{15}(x) = \sum_{i=1}^{11}\left[a_i - \frac{x_1(b_i^2 + b_i x_2)}{b_i^2 + b_i x_3 + x_4}\right]^2$ | 4 | $[-5, 5]$ | 0.0003 |
| $F_{16}(x) = 4x_1^2 - 2.1x_1^4 + \frac{1}{3}x_1^6 + x_1 x_2 - 4x_2^2 + 4x_2^4$ | 2 | [–5,5] | –1.0316 |

**Table A3.** Fixed-dimension multimodal benchmark functions[8].

| Function | Dimension | Range | $f_{min}$ |
|---|---|---|---|
| $F_{17}(x) = \left(x_2 - \frac{5.1}{4\pi^2}x_1^2 + \frac{5}{\pi}x_1 - 6\right)^2 + 10\left(1 - \frac{1}{8\pi}\right)\cos x_1 + 10$ | 2 | $[-5, 5]$ | 0.398 |
| $F_{18}(x) = [1 + (x_1 + x_2 + 1)^2(19 - 14x_1 + 3x_1^2 - 14x_2 + 6x_1x_2 + 3x_2^2)] \times [30 + (2x_1 - 3x_2)^2 \times (18 - 32x_1 + 12x_1^2 + 48x_2 - 36x_1x_2 + 27x_2^2)]$ | 2 | $[-2, 2]$ | 3 |
| $F_{19}(x) = -\sum_{i=1}^{4} c_i exp\left(-\sum_{j=1}^{3} a_{ij}\left(x_j - p_{ij}\right)^2\right)$ | 3 | [1,3] | –3.86 |
| $F_{20}(x) = -\sum_{i=1}^{4} c_i exp\left(-\sum_{j=1}^{6} a_{ij}\left(x_j - p_{ij}\right)^2\right)$ | 6 | [0,1] | –3.32 |
| $F_{21}(x) = -\sum_{i=1}^{5}[(X - \alpha_i)(X - \alpha_i)^T + C_i]^{-1}$ | 4 | [0,10] | –10.1532 |
| $F_{22}(x) = -\sum_{i=1}^{7}[(X-\alpha_i)(X-\alpha_i)^T + C_i]^{-1}$ | 4 | [0,10] | −10.4028 |
| $F_{23}(x) = -\sum_{i=1}^{1}0[(X-\alpha_i)(X-\alpha_i)^T + C_i]^{-1}$ | 4 | [0,10] | −10.536 |